\documentclass[letterpaper,twocolumn,10pt]{article}
\usepackage{usenix2019_v3}

\usepackage{tikz}
\usepackage{amsmath}

\usepackage{filecontents}

\usepackage{graphicx}
\usepackage{booktabs} 

\usepackage[labelfont=bf,font=small]{caption}
\usepackage{xspace}
\usepackage{subcaption}
\usepackage{comment}
\usepackage{listings}
\usepackage{hyperref}
\usepackage{enumitem}
\usepackage[small,compact]{titlesec}

\newcommand{\specialcell}[2][c]{%
  \begin{tabular}[#1]{@{}c@{}}#2\end{tabular}}

\newcommand{\minihead}[1]{{\vspace{.45em}\noindent\textbf{#1.} }}

\newcommand{\dawnbench}{\textsc{DAWNBench}\xspace}
\newcommand{\mlperf}{\textsc{MLPerf}\xspace}

\begin{document}

\date{}

\title{\Large \bf Analysis of DAWNBench, a Time-to-Accuracy Machine Learning
Performance Benchmark}

\author{
    {\rm Cody Coleman\thanks{Equal Contribution}, Daniel Kang$^*$, Deepak Narayanan$^*$, Luigi Nardi, Tian Zhao, Jian Zhang,} \\
    {\rm Peter Bailis, Kunle Olukotun, Chris R\'e, Matei Zaharia} \\
    {\rm Stanford DAWN}
}

\maketitle

\begin{abstract}

Researchers have proposed hardware, software, and algorithmic optimizations
to improve the computational performance of deep learning.
While some of these optimizations perform the
same operations faster (e.g., increasing GPU clock speed), many
others modify the semantics of the training procedure (e.g., reduced precision),
and can impact the final model's accuracy on unseen data.
Due to a lack of standard evaluation criteria that considers these trade-offs,
it is difficult to directly compare these optimizations.
To address this problem, we recently introduced \dawnbench, a benchmark
competition focused on \emph{end-to-end} training time to achieve
near-state-of-the-art accuracy on an unseen
dataset---a combined metric called time-to-accuracy (TTA).
In this work, we analyze the entries from \dawnbench, which received
optimized submissions from multiple industrial groups, to investigate the behavior
of TTA as a metric as well as trends in the best-performing entries.
We show that TTA has a low coefficient of variation and that
models optimized for TTA generalize nearly as well as those trained using standard methods.
Additionally, even though \dawnbench entries were able to train ImageNet models
in under 3 minutes, we find they still underutilize hardware capabilities such
as Tensor Cores. Furthermore, we find that distributed entries can spend more than
half of their time on communication.
We show similar findings with entries to the \mlperf v0.5 benchmark.

\end{abstract}

\section{Introduction}
Machine learning (ML) training has become an increasingly expensive computational workload.
In particular, deep learning (DL) enables users to train high-capacity models
with billions of parameters~\cite{baevski2018adaptive, jozefowicz2016exploring, chelba2013one}
from massive datasets that improve
in accuracy as the dataset grows~\cite{chen2017revisitingdata,amodei2018ai}.
Because modern DL methods are computationally expensive, researchers have proposed many hardware, software, and
algorithmic optimizations for DL, ranging from new
hardware platforms~\cite{burger2017microsoft, jouppi2017datacenter, pena2017benchmarking} and
software systems~\cite{abadi2016tensorflow, chetlur2014cudnn, chilimbi2014project, dean2012large, jia2014caffe}
to novel distributed optimization algorithms~\cite{de2017understanding, harlap2016addressing, recht2011hogwild,
zhang2014dimmwitted, glorot2011deep,
goyal2017accurate, iandola2016squeezenet, kingma2014adam, sohl2014fast,
sutskever2013importance}.

Unfortunately, performance evaluation for ML training systems is significantly more
challenging than performance evaluation for traditional software.
The main goal of ML training is to build a statistical model that \emph{generalizes}
well to new data, i.e., makes accurate predictions on it,
but many techniques that increase throughput can adversely affect generalization.
On the hardware side, large minibatch training~\cite{goyal2017accurate,
jouppi2017datacenter} and reduced precision~\cite{chilimbi2014project,
de2017understanding, micikevicius2017mixed} can help run iterations of the optimization algorithm faster and
speed up ``proxy'' metrics such as time to process an epoch (``time-per-epoch''),
but can prevent models from reaching the same accuracy on unseen data~\cite{masters2018revisiting, de2018high, mccandlish2018empirical}.
On the algorithmic side, techniques such as
the Adam optimizer~\cite{kingma2014adam} were shown to accelerate the
minimization of training loss (``time-to-training-loss'') but sometimes lead to
models with lower accuracy on unseen data~\cite{wilson2017marginal}.
These proxy metrics do not consider runtime and final model accuracy jointly,
making it hard to evaluate proposed computational optimizations.

To address this lack of standard evaluation criteria, we ran the
\dawnbench~\cite{coleman2017dawnbench} competition in 2018 to measure the end-to-end
performance of ML systems using a \emph{time-to-accuracy (TTA)} metric.
TTA measures time for a system to train to a target,
near-state-of-the-art accuracy level on a held-out dataset.
Unlike prior work that focused solely on throughput
metrics such as time-per-epoch~\cite{bahrampour2015comparative,
baidu2017deepbench, chintala2017convnet, adolf2016fathom, google2017benchmarks,
shi2016benchmarking}, TTA combines both generalization and speed.
While several papers had previously used TTA for evaluation~\cite{akiba2017extremely, goyal2017accurate, li2014scaling, smith2017learningratedecay},
\dawnbench was the first multi-entrant benchmark competition to use the TTA metric.
During the initial competition that ran in April 2018, Google,
Intel, fast.ai, and others submitted optimized entries that
could train to 93\% top-5 accuracy on ImageNet in less than 30 minutes, which subsequently dropped to under 3 minutes with rolling submissions.
Later that year, the \mlperf~\cite{mlperf2018} benchmark launched using TTA as its primary metric as well.

Despite the impressive speedups achieved by \dawnbench and \mlperf entries,
many questions remain about the performance of ML training systems and TTA as a metric.
For example, is the TTA metric stable or do the entries to these metrics only represent the
best result out of many trials?
Do models optimized for TTA still generalize well or are they implicitly adapting to
the held-out dataset used in the benchmark through extensive hyperparameter tuning?
Finally, how close are these entries from fully utilizing hardware platforms and
what are the computational bottlenecks?

In this paper, we evaluate entries from \dawnbench and from \mlperf v0.5 to
understand the behavior of TTA as an ML performance metric and identify
bottlenecks in the best performing entries. Both benchmarks received
professionally optimized entries from leading industry groups, such as the
Google TPU team, Intel, and NVIDIA, creating one of the first opportunities to
study ML systems optimized heavily for \emph{training performance}, as opposed
to traditional ML competitions that only evaluate
accuracy~\cite{imagenet_cvpr09}.  Fortunately, most of the top entries were open
source. Using these top-performing, open-source benchmark entries, we find
that:
\vspace{-0.5em}
\begin{enumerate}[leftmargin=1.25em,itemsep=-0.2em]
  \item Despite the stochasticity of ML training procedures, TTA is
  a relatively stable metric that can reliably distinguish between systems on tasks
  that include image classification, object detection, and machine translation (\S\ref{sec:variability}).

  \item Even though accuracy in TTA is measured on a fixed, held-out evaluation set,
  models optimized for TTA generalize to unseen data nearly as well
  as off-the-shelf models (\S\ref{sec:generalization}).

  \item Distributed training often bottlenecks on communication (often $> 50\%$ of
  total time spent on communication), both on publicly available cloud
  infrastructure and optimized on-premise deployments with fast networks
  (\S\ref{sec:distributed-training}).

  \item Some of the top-performing benchmark entries \emph{severely} underutilize hardware capabilities such
  as Tensor Cores by up to 10$\times$.
  
  \item Training is bottlenecked by operators previously thought to be
  inexpensive, such as rectified linear units (ReLUs)~\cite{nair2010rectified} (\S\ref{sec:hw-single-node}).
\end{enumerate}

\section{Background: ML Training}
\label{sec:ml}

In this section, we describe the ML training workload and how it differs in performance
goals from other applications.

\begin{figure}
  \includegraphics[width=0.9\columnwidth]{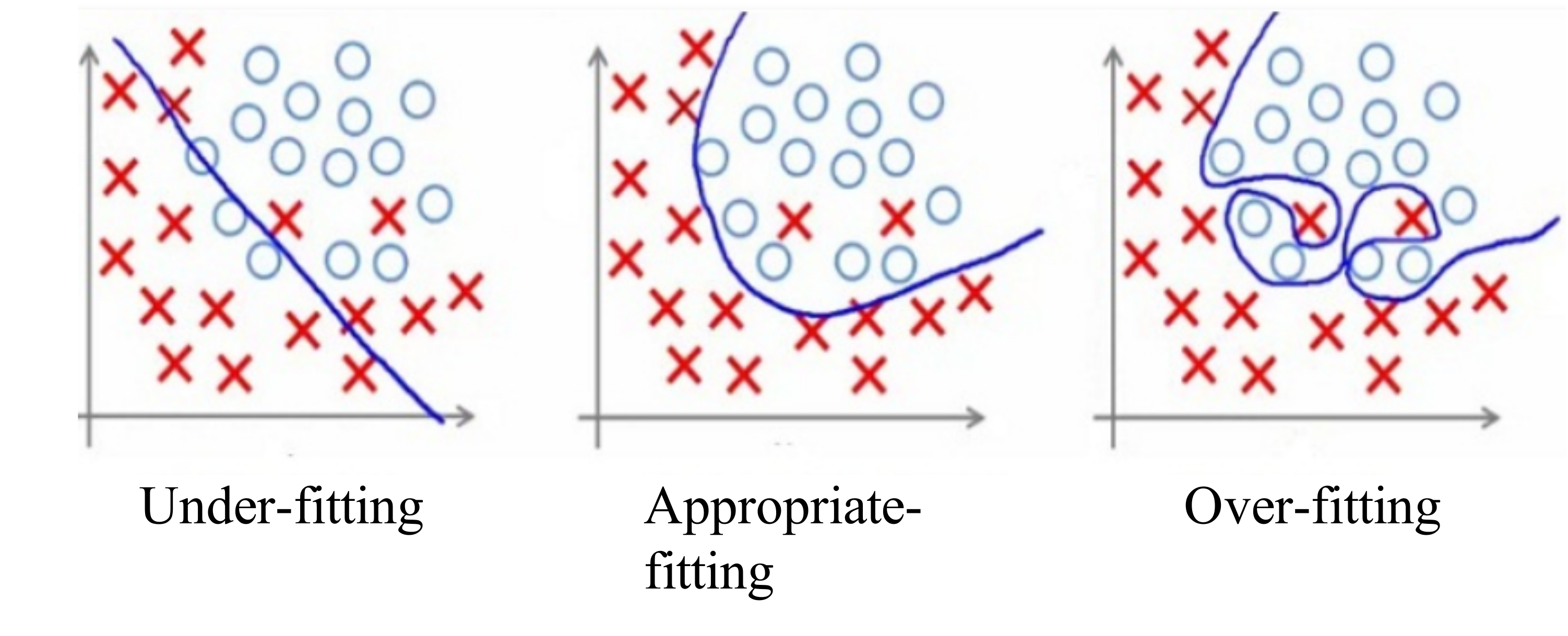}
  \vspace{-0.5em}
  \caption{Examples of underfitting, appropriate fitting, and overfitting for ML
  models. The overfit model classifies the training data perfectly, but will
  perform worse on unseen data than the middle model. Figure
  adapted from~\cite{anup2018what}.}
  \label{fig:overfitting}
\end{figure}

\minihead{The Goal of ML Training: Generalization}
The main goal of ML is to train a model that makes high quality predictions on unseen data, which is referred to as
\emph{generalization}~\cite{goodfellow2016deep}.
An optimization algorithm minimizes a problem-specific loss function to find a model
that not only performs well on the training data, but is also likely to \emph{generalize}
to unseen data from a similar distribution.
This goal is different from pure mathematical optimization, as shown in Figure~\ref{fig:overfitting}:
for example, when the ML algorithm can propose a large range of functions as models, it is
possible to \emph{overfit} the training data and return a model that generalizes less well to
unseen data than a simpler model.
Deep learning models in particular have the capacity to represent a very wide range of
functions, so much of DL research focuses on finding methods that generalize well~\cite{goodfellow2016deep}.

\begin{table*}[t!]
\begin{subtable}[t]{0.48\textwidth}
  \centering
  \setlength\itemsep{2em}
  \small
  \begin{tabular}{ccc}
    \specialcell{Hardware}
      & \specialcell{\# of entries}  \\ \hline \hline
    GPU      & 8   \\
    TPU      & 8   \\
    CPU      & 3   \\
    \\
  \end{tabular}
  \centering
  \setlength\itemsep{2em}
  \small
  \begin{tabular}{ccc}
    \specialcell{Framework}
      & \specialcell{\# of entries}  \\ \hline \hline
    TensorFlow  & 11    \\
    PyTorch     & 4     \\
    Caffe       & 3     \\
    MXNet       & 1    
  \end{tabular}
  \caption{Overview of hardware platforms and software frameworks for
  \dawnbench ImageNet submissions.}
  \label{table:dawnbench-entry-stats}
\end{subtable}
\hspace{0.5em}
\begin{subtable}[t]{0.48\textwidth}
  \centering
  \setlength\itemsep{2em}
  \small
  \begin{tabular}{ccc}
    \specialcell{Hardware}
      & \specialcell{\# of entries}  \\ \hline \hline
    GPU      & 28  \\
    TPU      & 7   \\
    CPU      & 6   \\
    \\
    \\
  \end{tabular}
  \centering
  \setlength\itemsep{2em}
  \small
  \begin{tabular}{ccc}
    \specialcell{Framework}
      & \specialcell{\# of entries}  \\ \hline \hline
    PyTorch     & 22   \\
    TensorFlow  & 10   \\
    Caffe       & 3    \\
    MXNet       & 5    \\
    Big DL      & 1
  \end{tabular}
  \vspace{-0.5em}
  \caption{Overview of the hardware platforms and software frameworks for
  \mlperf entries. We excluded ``research'' submissions, which include frameworks
  and hardware not publicly available.}
  \label{table:dawnbench-entry-stats}
\end{subtable}
\vspace{-0.5em}
\caption{Summary of infrastructure used for \dawnbench and \mlperf entries.}
\label{table:entry-stats}
\end{table*}

\begin{table*}
\centering
\small
\setlength\itemsep{2em}
\begin{tabular}{llllll}
  Model       & Area     & Problem & Dataset & Dataset size & Quality target \\
  \hline \hline

  ResNet      & Vision   & Image classification & ImageNet & 1.2M images & 74.9\% \\
  SSD, ResNet-34 backbone
              & Vision   & Object detection     & MS-COCO & 127K images & 21.1 mAP \\

  Mask R-CNN, & Vision   & Object detection and & MS-COCO & 127K images & 37.7 box mAP, \\
  \xspace \xspace ResNet-50 backbone &   & \xspace \xspace instance segmentation
  & & & \xspace \xspace 33.9 mask mAP \\

  \hline
  GNMT        & Language & Translation (recurrent) & WMT English-German & 3.5M sent. & 21.8 BLEU \\
  Transformer & Language & Translation (non-recurrent) & WMT English-German & 4.6M sent. &  25.0 BLEU  \\
  \hline
  NCF         & Commerce & Recommendation & MovieLens-20M  & 20M ratings & 0.635 HR@10 \\
  \hline
  Minigo      & RL & Go & Go & Self-play & 40.00\% accuracy
\end{tabular}
\caption{Overview of tasks, models, and problem areas for the \mlperf v0.5
training benchmark.}
\label{table:mlperf-tasks}
\end{table*}

To quantify how well a model generalizes, a separate dataset is held out
from training and used for periodic evaluation. This dataset is referred to the
validation dataset.
To avoid overfitting, most systems stop training when performance on the validation
set has plateaued.\footnote{
  Some texts also use the term ``validation set'' to refer to data held out for hyperparameter
  tuning, and use ``test set'' for evaluation data.
}

However, even with a held-out validation set, repeated experiments could lead to
overfitting, even though the model was never explicitly trained on the held-out data~\cite{recht18cifar}.
While tuning hyperparameters to optimize for TTA, entries could implicitly be
learning about and adapting to the validation set rather than achieving the principal goal of generalization.
Fortunately, this form of overfitting does not seem to occur in the existing \dawnbench and \mlperf entries (\S~\ref{sec:generalization}).

\minihead{Typical Training Processes}
Most deep learning models are trained using Stochastic Gradient Descent~\cite{robbins1951stochastic}
or one of its accelerated variants, such as Adam~\cite{kingma2014adam}.
These methods iterate over the training data in \emph{minibatches},
which are small batches of records (e.g., 32 records) drawn at random.
The training algorithm updates the weights of the model after processing each batch.
In total, the optimization method may make multiple passes over the entire
dataset during training, where each complete pass is called an \emph{epoch}.

\minihead{Tradeoffs in Speed and Generalization}
Unlike more traditional workloads, many optimizations that improve how fast the ML system processes data affect the quality of the solution, either changing how many updates it takes for the model to converge or preventing the model from converging to the same quality.
For example:
\vspace{-0.5em}
\begin{enumerate}[leftmargin=1.25em,itemsep=-0.2em]
  \item Increasing the number of records used for each update can increase
  hardware efficiency, but prevent or slow down
  convergence~\cite{mccandlish2018empirical}
  (\S~\ref{sec:metric-comparison}).

  \item Naively reducing floating point precision to 16 bits prevents
  convergence, but using ``loss scaling'' allows for
  convergence~\cite{micikevicius2017mixed}. Further reducing to 8 bits
  generally prevents convergence with current methods~\cite{de2018high}.

  \item In the multi-accelerator case, SGD can be performed synchronously or asynchronously~\cite{recht2011hogwild}. Synchronicity ensures that each update uses the most up-to-date weights of the model to accurately assess performance, but requires more overhead to copy the model's weights between accelerators after each update. Asynchronous SGD can remove this synchronization at the cost of data efficiency~\cite{li2014scaling,mitliagkas2016asynchrony}.
\end{enumerate}

\minihead{Stochasticity in Training}
Training via SGD is inherently stochastic. Stochasticity enters in several ways,
including randomness in model initialization and data traversal.
Furthermore, many DL systems introduce stochasticity for improved hardware
efficiency, e.g., by reordering floating point operations.
Thus, multiple trials of the same optimization procedure can reach
the same target validation accuracy in a different number of epochs.

\section{Overview of Benchmarks}
\label{sec:benchmark}

This section overviews the rules, training procedures, and models from \dawnbench and \mlperf.
We also detail the entries we leverage in our subsequent analysis.

\subsection{\dawnbench Overview}
\dawnbench was introduced in November 2017 and concluded in April 2018.
\dawnbench evaluates the time and cost (in USD) of popular deep learning
training and inference workloads. The initial release included two tasks: image
classification on ImageNet and CIFAR10, and question answering on SQuAD, and
four metrics: training time to a specified validation accuracy, cost of training
to that accuracy for submissions that use hardware in the public cloud, average
latency of performing inference on a single item (image or question), and
average inference cost.

Entries were required to submit a description of their submission and the
validation accuracy after every epoch. While source code was optional, every
submission for image classification included a link to all code needed to
reproduce runs, assuming access to the appropriate hardware. For question
answering on SQuAD, some submissions did not include code until well after the
\dawnbench deadline; because of this and the general lack of submissions, we
focus exclusively on image classification \emph{training} submissions in this
paper. While our analysis applies to both ImageNet and CIFAR10, we do not include
results for CIFAR10 in this paper since CIFAR10 does not reflect the scale of production workloads.
As a result our analysis of \dawnbench focuses solely on ImageNet, where \dawnbench used a top-5 accuracy target of 93\%.

\subsection{\mlperf Overview}
\mlperf v0.5 is a more recent benchmark that concluded in December 2018. \mlperf evaluates TTA on a broader range of tasks, including
image classification, object detection, translation, and recommendation, as shown in Table~\ref{table:mlperf-tasks}.
Unlike \dawnbench, \mlperf used a fixed model and optimization algorithm.  There was some
flexibility for choosing SGD hyperparameters to allow
submissions of different computational scales.
Submissions were also allowed to submit results for a subset of tasks, so the
majority of hardware targets did not include entries for every task.
For example, the reinforcement learning task had no entries with accelerators, as game simulation was the bottleneck.
As such, we do not analyze the reinforcement learning entries.
Similarly, we do not analyze the results on the recommendation task because
it does not reflect production usage and will be replaced~\cite{bittorf2019making}.

\subsection{Summary of Entries}
Entries to \dawnbench and \mlperf v0.5 came from many organizations, including Google, NVIDIA, and Intel, which had teams
of engineers optimize their submissions. The entries spanned GPUs, TPUs, and
CPUs on the hardware side and TensorFlow~\cite{abadi2016tensorflow}, PyTorch~\cite{paszke2017automatic},
Caffe~\cite{jia2014caffe}, MXNet~\cite{chen2015mxnet}, and Big DL~\cite{intel2019bigdl} on the
software side. The number of compute units (which we refer to as compute scale) ranged from 2 to 640
processors, and speedups over reference implementations ranged from 1.6$\times$ to
over 1,400$\times$.
In \mlperf v0.5, every entry with an accelerator used mixed precision
training~\cite{micikevicius2017mixed}, and large batch sizes~\cite{goyal2017accurate}. 
\dawnbench submissions were allowed to use a wider range of optimizations, including progressive resizing of
images~\cite{lim2017enhanced, karras2017progressive} and novel model
architectures~\cite{real2018regularized}, in addition to mixed-precision training and large minibatch training.
In our analysis, we used all pre-February 2019 submissions that were reproducible
with public cloud infrastructure or included sufficient information for analysis (e.g., training logs).

\section{Analysis of Time-to-Accuracy}
\label{sec:metric_eval}

In this section, we evaluate the TTA metric along three axes, using
publicly available code and results from \dawnbench and \mlperf submissions. First, we demonstrate that TTA
has a low coefficient of variation ($<14\%$) over several runs with
fixed hyperparameters, even with some statistical optimizations (e.g., cyclic
learning rates, progressive resizing) that result in higher variance.
Second, we provide evidence that models optimized for TTA generalize nearly as well
as regular, unoptimized models. Third, we
compare TTA against other metrics and show that the alternative
metrics do not capture the complexity of DL training.

\subsection{Variability of Time-to-Accuracy}
\label{sec:variability}

\begin{table*}[t]
\begin{subtable}[t]{0.48\textwidth}
\centering
\setlength\itemsep{2em}
\small
\begin{tabular}{lccc}
  \specialcell{Entry name} & \specialcell{Coeff. of variation} & \specialcell{Frac. of runs} \\ \hline \hline
  ResNet-50, \texttt{p3.16xlarge}      & 5.3\%                    & 80\% \\
  ResNet-50, 4x\texttt{p3.16xlarge}     & 11.2\%                   & 60\% \\
  ResNet-50, 8x\texttt{p3.16xlarge}     & 9.2\%                    & 100\% \\
  ResNet-50, 16x\texttt{p3.16xlarge}    & 12.2\%                   & 100\% \\
  ResNet-50, 1xTPU       & 4.5\%                    & 100\% \\
  AmoebaNet-D, 1xTPU     & 2.3\%                    & 100\% \\
  ResNet-50, 1/2 TPU Pod & 2.5\%                    & 100\%
\end{tabular}
\caption{Coefficient of variation and fraction of runs that reached the desired
target accuracy of the top \dawnbench entries for image classification on ImageNet (5
runs). \texttt{p3.16xlarge} entries were from fast.ai and used progressive resizing. We also include the
coefficient over 4 runs of 1/2 a TPU Pod for ResNet-50.}
\label{table:dawnbench-entry-variance}
\end{subtable}
\hspace{0.5em}
\begin{subtable}[t]{0.48\textwidth}
\centering
\setlength\itemsep{2em}
\small
\begin{tabular}{lc}
  \specialcell{Entry} & \specialcell{Coeff. of variation} \\ \hline \hline
  ResNet, NVIDIA, 1xDGX-1  &  6.7\% \\
  SSD, NVIDIA, 1xDGX-1     &  0.5\% \\
  SSD, NVIDIA, 8xDGX-1     &  6.7\% \\
  Mask, NVIDIA, 1xDGX-1    &  3.9\% \\
  Mask, NVIDIA, 8xDGX-1    &  0.8\% \\
  GNMT, NVIDIA, 1xDGX-1    &  0.2\% \\
  Transformer, NVIDIA, 1xDGX-1 & 13.8\%
\end{tabular}
\caption{Coefficient of variation for selected official \mlperf entries. All of
the displayed runs achieved the target accuracy 100\% of the time. We exclude
recommendation as the model and dataset are being replaced for the next version of \mlperf.}
\label{table:mlperf-entry-variance}
\end{subtable}
\vspace{-0.5em}
\caption{Coefficient of variation and fraction of runs that achieved the target
accuracy for \mlperf and \dawnbench entries. As shown, the coefficient of
variation is less than 14\% for all runs, with the exception of multi-accelerator
Transformer entries (not shown). However, \mlperf plans to expand the dataset
size for Transformer, which we believe will improve stability.}
\label{table:entry-variance}
\end{table*}

To understand the stability of TTA, we computed the coefficient of variation (the ratio of the variance to the mean) for the top \dawnbench entries
available on public cloud (by rerunning them several times) and \emph{official} \mlperf
entries (which contained multiple trials). We chose this metric
as the mean is a natural scale for comparing systems. For example, a coefficient
of variation of 14\% means that systems that achieve a TTA within
14\% of each other are not easily distinguished, but a system that is two times
faster than another is easy to distinguish.

As shown in Table~\ref{table:dawnbench-entry-variance}, the
coefficient of variation of TTA for the reproduced \dawnbench entries is at most 4.5\% for entries that
do not use novel statistical optimizations, but 12.2\% for all entries. This
indicates that TTA is largely stable despite the randomness in
DL training.

We also found that several entries failed to consistently achieve the given accuracy
threshold. In particular, progressive resizing used by several of the
\dawnbench ImageNet entries appear to make validation convergence less robust
as seen in Table~\ref{table:dawnbench-entry-variance}.

The coefficient of variation was similarly low for the official \mlperf results.
 Table~\ref{table:mlperf-entry-variance} shows the coefficient of variation for the official \mlperf results.
We find that TTA is largely stable;
the coefficient of variation is always less than 14\% and generally less than 7\%.
We additionally reproduced the majority of
available \mlperf entries on stable public cloud hardware. We found
that these reproduced \mlperf entries were in line with the official entries.

\begin{figure*}[t!]
\centering
\begin{subfigure}{0.65\columnwidth}
  \includegraphics[width=1.0\columnwidth]{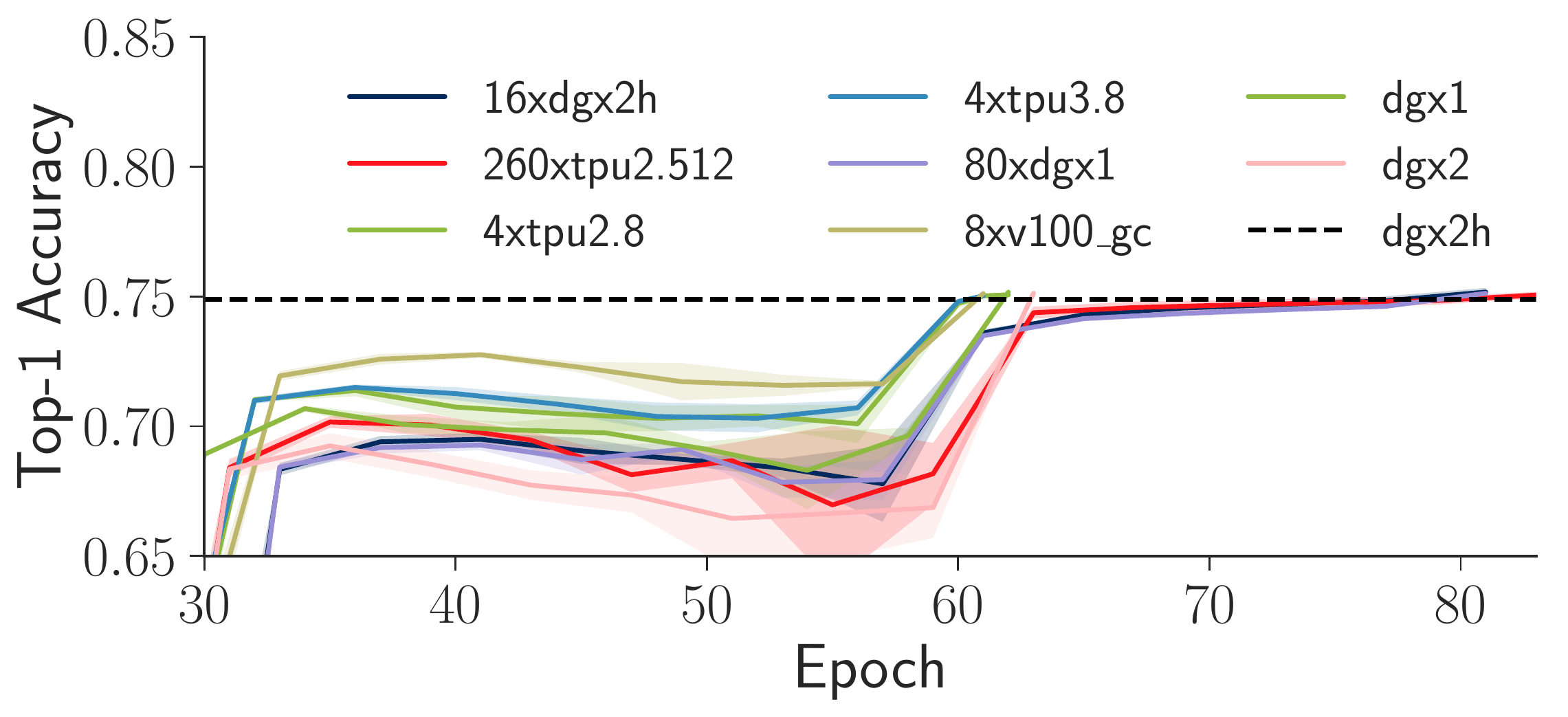}
  \caption{ResNet}
\end{subfigure}
\begin{subfigure}{0.65\columnwidth}
  \includegraphics[width=1.0\columnwidth]{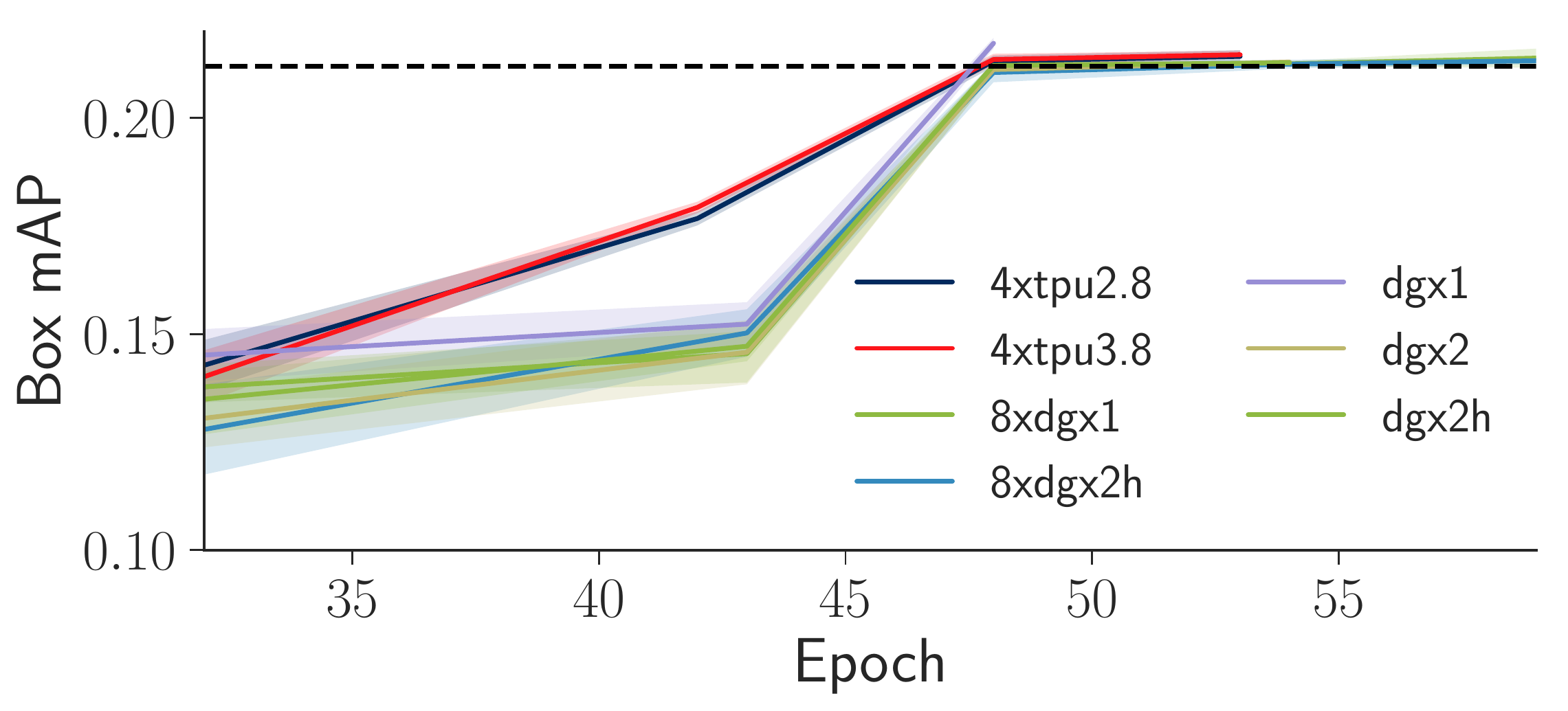}
  \caption{SSD}
\end{subfigure}
\begin{subfigure}{0.65\columnwidth}
  \includegraphics[width=1.0\columnwidth]{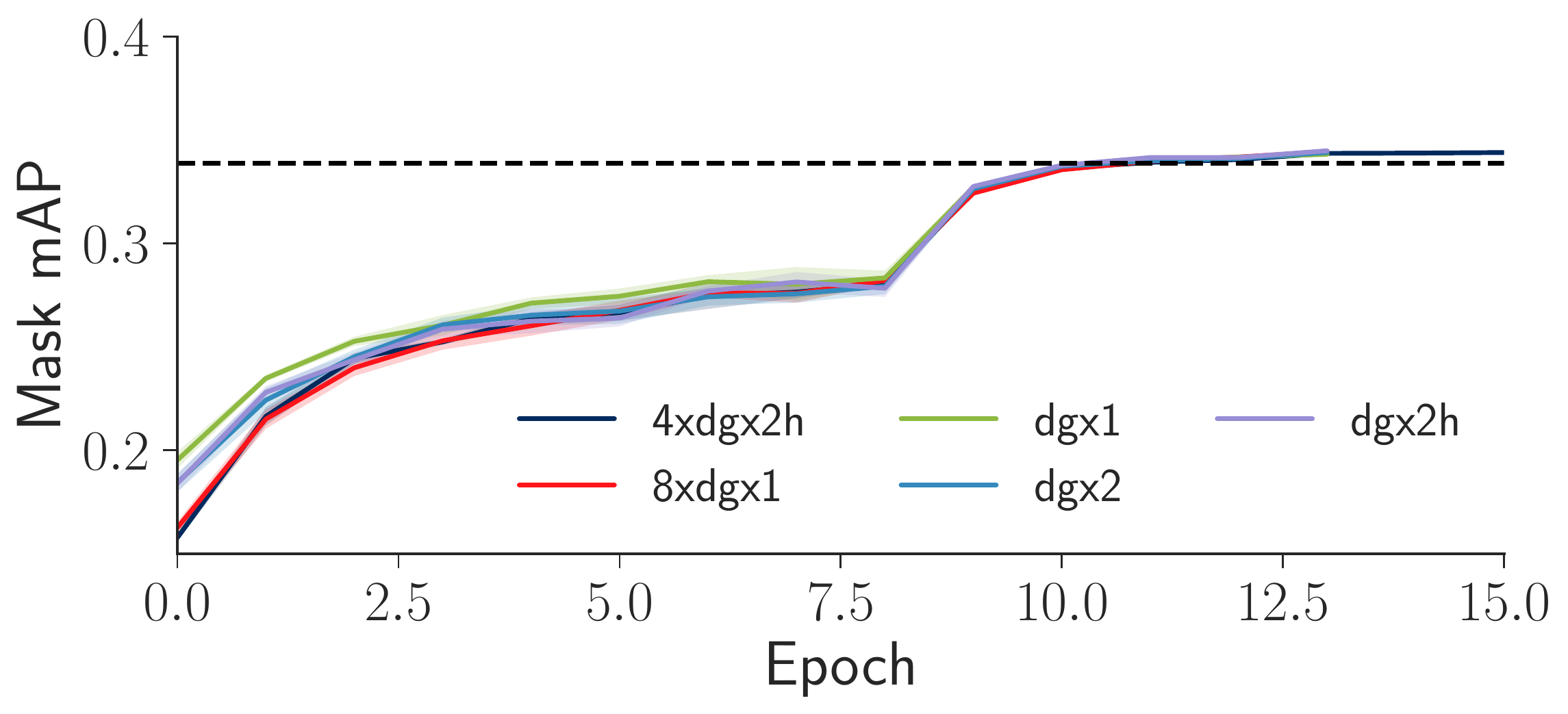}
  \caption{Mask}
\end{subfigure}
\begin{subfigure}{0.65\columnwidth}
  \includegraphics[width=1.0\columnwidth]{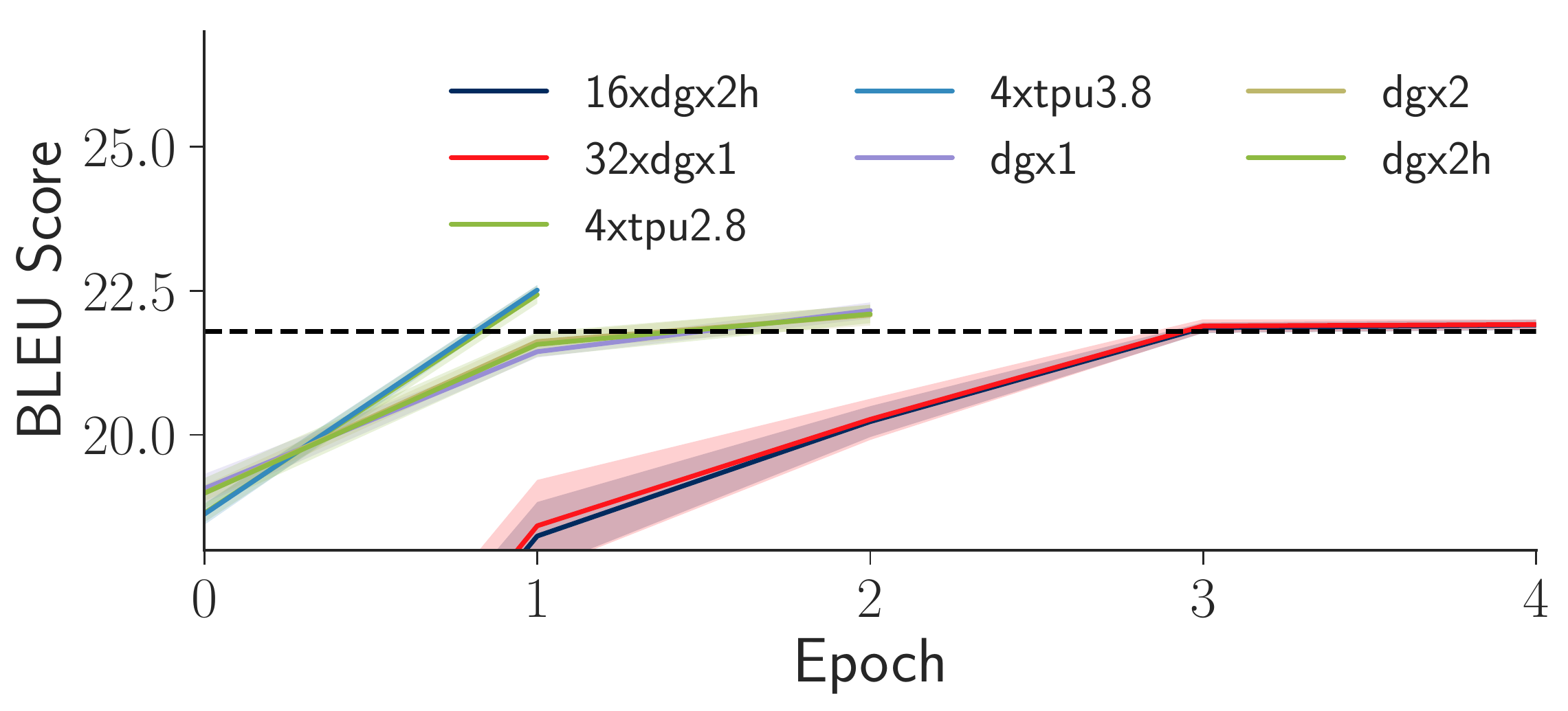}
  \caption{GNMT}
\end{subfigure}
\begin{subfigure}{0.65\columnwidth}
  \includegraphics[width=1.0\columnwidth]{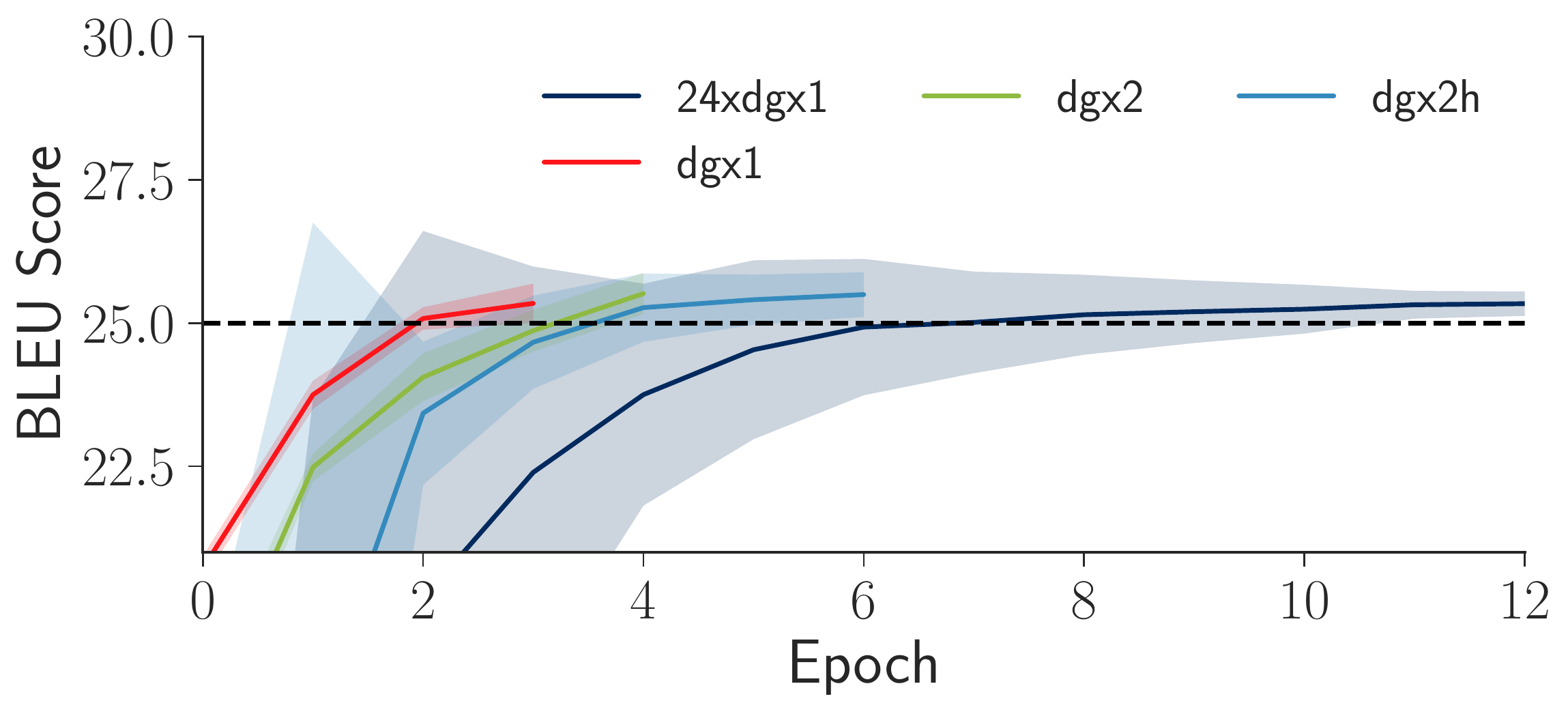}
  \caption{Transformer}
\end{subfigure}
\vspace{-0.5em}
\caption{Variation in validation quality metric per epoch for official \mlperf
entries. As shown, the variation decreases closer to the final target
score. The variation within an epoch is typically smaller than the variation
between epochs close to the target score. Best seen in color.}
\label{fig:mlperf-epoch-var}
\end{figure*}

\minihead{Source of Variation}
To understand the source of variation in TTA, we analyzed the validation
accuracy curves per epoch for \mlperf entries.
Figure~\ref{fig:mlperf-epoch-var} shows the variance in quality metric per epoch
across several tasks and machine scales.
Validation accuracy is less stable at the beginning of training but becomes more stable as training continues.
This variance early in training grows with the system scale because larger entries start training with large learning rates.
Additionally, the variation in the number of epochs is high due to the different machine scales.
For selected large scale entries, Table~\ref{table:source-of-var} shows low variation in time-per-epoch,
with a coefficient of variation less than 3\%. The variation in
the number of epochs to reach the target quality metric is up to 45$\times$ higher than the variation in time-per-
epoch. Thus, most of the variation in TTA comes from variation in the number
of epochs.

\subsection{Generalization of Optimized Models}
\label{sec:generalization}

To measure the generalization performance of models optimized for
TTA in image classification and translation,
we collect unseen data, i.e., data that is not in the validation and training
sets, and test the accuracy on this unseen data.
We used reproduced \dawnbench and \mlperf entries since
neither benchmark provided checkpoints.

\minihead{Evaluation on New Data for Image Classification}
To test image classification, we scraped and labeled a set of 2,864 images from
Flickr. The images were scraped based on the WordNet keywords associated with
each class in the ImageNet dataset. The top five images based on relevance were
shown to a human labeler and labeled correct or incorrect. To ensure no overlap
with ImageNet, only images posted after January 1st, 2014 were used. The images
spanned 886 (out of 1000) classes. While these images are not entirely
representative of ImageNet, we believe they reflect a reasonable distribution.

\begin{table}[t!]
\centering
\small
\begin{tabular}{lccc}
  Entry  &  Epochs  &  \specialcell{Time per\\epoch (s)}  &
  \specialcell{Ratio of\\std. devs} \\ \hline \hline
  ResNet, 80xDGX-1  &  $82 \pm 0$     &  $4.6 \pm 0.01$   & 0$\times$ \\
  SSD, 8xDGX-2      &  $53.6 \pm 4.2$ &  $6.3 \pm 0.03$   & 14.8$\times$ \\
  Mask, 4xDGX-2H    &  $14.4 \pm 0.8$ &  $310.0 \pm 2.9$  & 5.9$\times$ \\
  GNMT, 16xDGX-2H   &  $4.2 \pm 0.4 $ &  $39.2 \pm 1.0$   & 3.9$\times$ \\
  Transf., 24xDGX-1 &  $7.2 \pm 2.1 $ &  $56.2 \pm 0.4$   & 44.9$\times$
\end{tabular}
\caption{The deviation in number of epochs, time per epoch, and the ratio of the
standard deviations for selected official \mlperf entries. We selected the largest scale
entries. The variation largely comes
from the variation in the number of epochs, not the time per epoch.}
\label{table:source-of-var}
\end{table}

\begin{table}[t!]
\begin{subtable}[t]{0.48\textwidth}
  \centering
  \small
  \begin{tabular}{lcc}
    \specialcell{Model} & \specialcell{Accuracy (top-5, unseen data)} \\ \hline \hline
    ResNet-18 (pretrained)  & 89.5\% \\
    ResNet-50 (pretrained)  & 92.2\% \\
    ResNet-152 (pretrained) & 93.2\% \\
    \hline
    ResNet-50, 1xTPU       & 92.6\% \\
    ResNet-50, \texttt{p3.16xlarge}   & 91.9\% \\
    ResNet-50, 4x\texttt{p3.16xlarge}  & 91.3\% \\
    ResNet-50, 8x\texttt{p3.16xlarge}  & 91.5\% \\
    ResNet-50, 16x\texttt{p3.16xlarge} & 91.3\% \\
    AmoebaNet-D, 1xTPU       & 91.3\% \\
  \end{tabular}
  \caption{\dawnbench submissions, top-5 accuracy. ResNet-50 on
  \texttt{p3.16xlarge} instances used non-standard optimizations such as
  progressive resizing.}
\end{subtable}
\hspace{0.5em}
\begin{subtable}[t]{0.48\textwidth}
  \centering
  \small
  \begin{tabular}{lcc}
    \specialcell{Model} & \specialcell{Accuracy (top-1, unseen data)} \\ \hline \hline
    ResNet-18 (pretrained)       & 71.7\% \\
    ResNet-50 (pretrained)       & 77.4\% \\
    ResNet-152 (pretrained)      & 79.4\% \\
    \hline
    ResNet-50, DGX-1   & 77.6\%
  \end{tabular}
  \caption{\mlperf submission, top-1 accuracy.}
\end{subtable}
\vspace{-0.5em}
  \caption{Performance of pre-trained models and models optimized for
  TTA on unseen data for \dawnbench and \mlperf ImageNet entries.
  The models optimized for TTA perform nearly as well as or better
  than the PyTorch pre-trained model. We expect the pretrained ResNet-18 and
  ResNet-152 to be lower and upper bounds respectively on generalization
  performance.}
  \label{table:flickr}
\end{table}

We computed the relevant accuracy metric (top-1 or top-5 accuracy) for
\dawnbench entries, an optimized \mlperf entry, and pre-trained ResNet-50
weights provided by PyTorch on the images from Flickr. The results are
summarized in Table~\ref{table:flickr}. As shown, the models optimized for
TTA achieve nearly the same accuracy or higher than the pre-trained
ResNet-50, indicating that optimizing for TTA does not sacrifice
generalization performance.

\minihead{Evaluation on Unseen Data for Translation Tasks}
For the \mlperf GNMT and Transformer models, we additionally used new data to
test generalization performance. We used the WMT'17 English-German
\texttt{newstest2017} test set~\cite{wmt2017}, which is not in the training or validation
sets for GNMT or Transformer.
Table~\ref{table:trans-generalization} shows the
optimized implementations generalize as well as the reference implementations,
despite being over 50$\times$ faster. This indicates that optimizing for
TTA does not sacrifice generalization performance.

\begin{table}
\centering
\setlength\itemsep{2em}
\small
\begin{tabular}{ll}
  Model             & BLEU score \\ \hline \hline
  GNMT, reference   & $23.44 \pm 0.08$ \\
  GNMT, DGX-1       & $23.63 \pm 0.20$ \\
  Transformer, reference & $26.60 \pm 0.44$ \\
  Transformer, DGX-1     & $26.78 \pm 0.45$ \\
\end{tabular}
\caption{BLEU scores on unseen data for the reference and optimized GNMT and
Transformer models. As shown, models optimized for TTA generalize
as well as the reference models. We show the average of three runs and the
standard deviation.}
\label{table:trans-generalization}
\end{table}

\subsection{Comparison to Alternative Metrics}
\label{sec:metric-comparison}
\minihead{Comparison to Throughput}
To demonstrate that throughput (and equivalently time-per-epoch and achieved FLOPS for a fixed model and dataset) is not sufficient for measuring DL system
performance, we show the batch size, number of epochs, throughput speedup, and
TTA speedup in Table~\ref{table:tta-vs-throughput}.
As shown, TTA speedups and throughput speeds can differ by up to 3$\times$, as
increasing the batch size to improve throughput can increase the number of epochs
required for convergence.
Thus, throughput is insufficient to characterize DL system performance even though it has been used extensively in prior benchmarks~\cite{baidu2017deepbench,
chintala2017convnet, google2017benchmarks, shi2016benchmarking}.

\begin{table}[t!]
\centering
\small
\setlength\itemsep{2em}
\begin{tabular}{lccccc}
  Model  & \specialcell{System\\scale} & BSes  & Epochs & \specialcell{Thpt.\\speedup} & \specialcell{TTA\\speedup}   \\ \hline \hline
  Trans. & 1, 24 & 10k, 492k  & 2, 6    & 10.9$\times$  &  3.6$\times$  \\
  GNMT   & 1, 32 & 1k, 8.2k   & 3, 5    & 10.9$\times$  &  6.5$\times$  \\
  ResNet & 1, 80 & 4k, 16k    & 63, 82  & 28.2$\times$  &  21.6$\times$ \\
  SSD    & 1, 8  & 1.2k, 2k   & 49, 55  & 4.6$\times$   &  4.1$\times$  \\
  \specialcell{Mask\\R-CNN} & 1, 8  & 32, 128    & 13, 14  & 4.2$\times$   &  3.9$\times$
\end{tabular}
\vspace{-0.5em}
\caption{Model, system scale (in number of DGX-1s), batch size (BS), number of
epochs for convergence, throughput speedup, and TTA speedup.
Numbers are given for two system scales per model using official \mlperf
entries. As shown, throughput does not directly correlate with TTA and speedups can
differ by up to 3$\times$ ($10.9\times$ vs $3.6\times$ for transformer).}
\label{table:tta-vs-throughput}
\end{table}

\minihead{Comparison to Peak Device FLOPS}
As shown in Table~\ref{table:tta-vs-throughput}, system scale does not correlate with
throughput or TTA speedup. We further show in
\S~\ref{sec:hw-single-node} that existing accelerators are severely underutilized in many cases. Additionally, other costs (e.g., communication overhead) can
dominate runtimes and can add a $> 2\times$ overhead. Thus, peak device
FLOPS is a poor proxy for observed DL system performance.

\section{Hardware Utilization and Scaling}
\label{sec:hw-util}

In this section, we evaluate how well highly optimized \dawnbench and \mlperf entries utilize available hardware.
First, we demonstrate that distributed entries can spend more than half of their time on communication overhead.
Second, we study the utilization of these entries on a single worker.
Through a roofline analysis~\cite{williams2009roofline}, we provide evidence that despite near state-of-the-art
training performance across a range of tasks, many submissions
still \emph{severely} underutilize the available hardware resources.
We also show that memory-bound kernels take a significant percentage of total runtime,
leading to lower observed FLOPS.

\begin{figure*}[t!]
  \centering
  \begin{subfigure}{.65\columnwidth}
    \includegraphics[width=1.0\columnwidth]{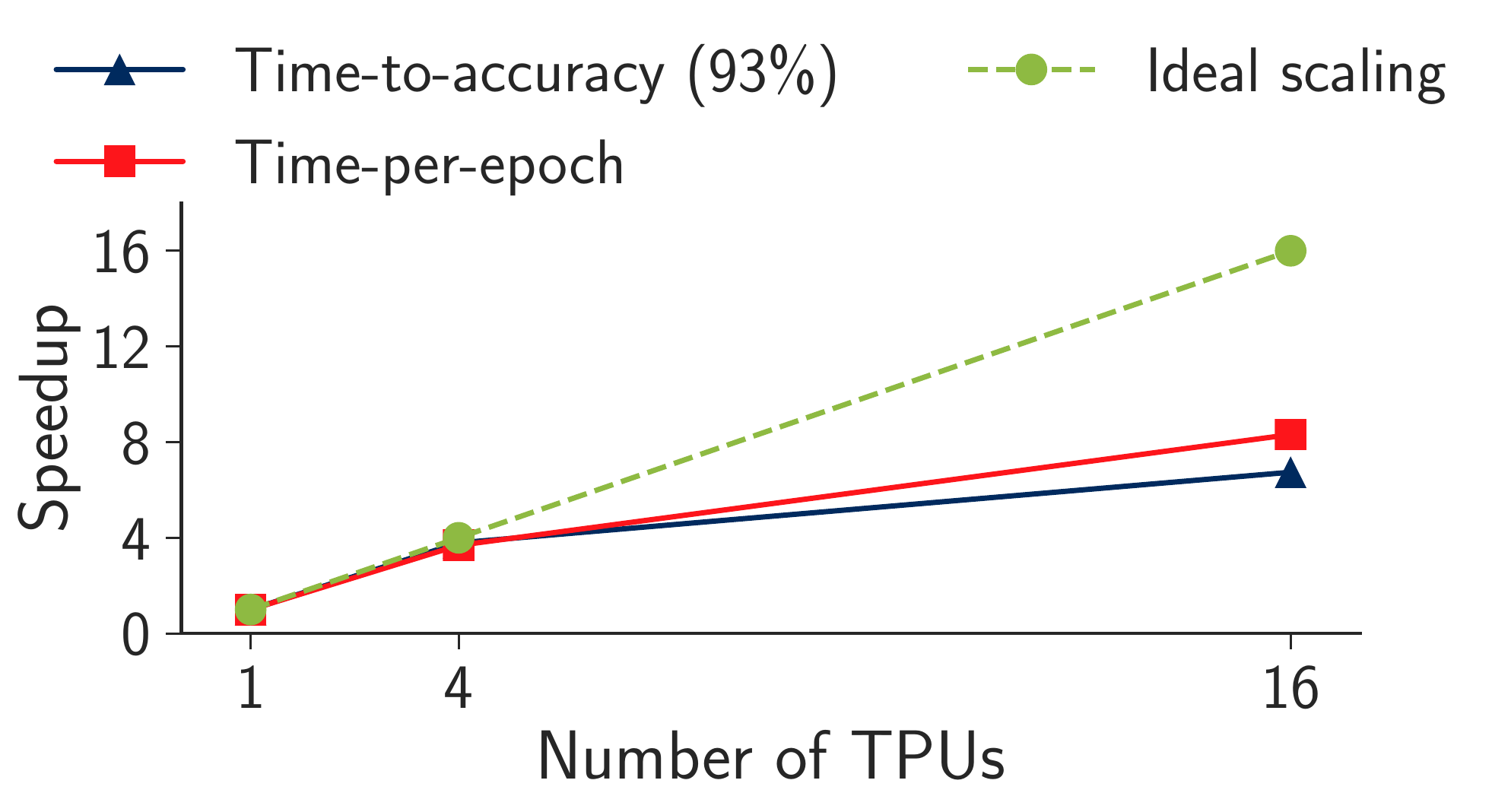}
    \caption{
      AmoebaNet across TPUs, TPU pod.
    }
    \label{fig:speedup-vs-num-workers-amoebanet}
  \end{subfigure}
  \begin{subfigure}{.65\columnwidth}
    \includegraphics[width=1.0\columnwidth]{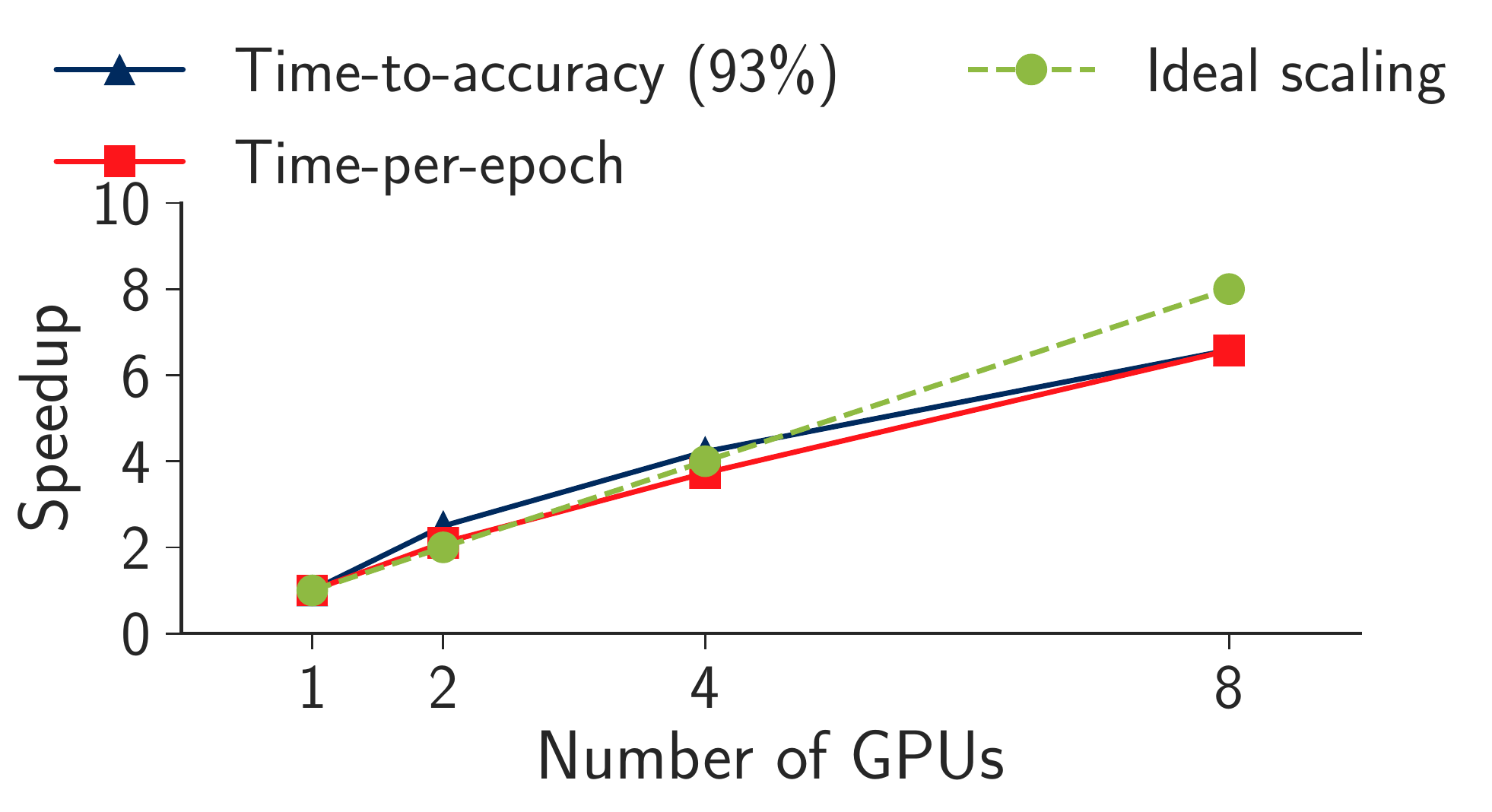}
    \caption{
      ResNet-50 within \texttt{p3.16xlarge} server.
    }
    \label{fig:speedup-vs-num-workers-resnet50}
  \end{subfigure}
  \begin{subfigure}{.65\columnwidth}
    \includegraphics[width=1.0\columnwidth]{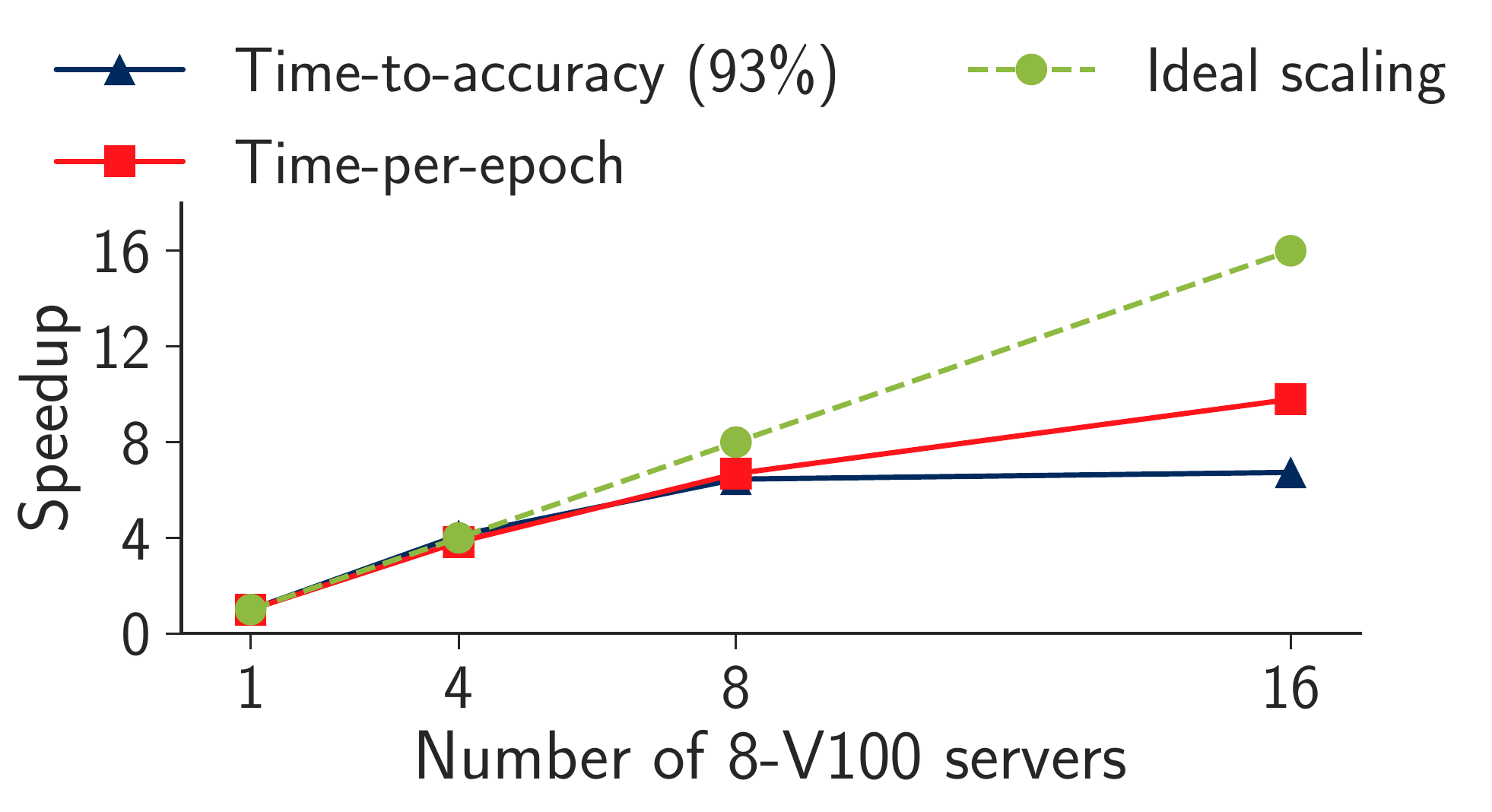}
    \caption{
      ResNet-50 across \texttt{p3.16xlarge} servers.
    }
    \label{fig:speedup-vs-num-workers-resnet50-dgx1}
  \end{subfigure}
  \caption{
    Speedup with respect to a single worker vs. number of workers for three
    ImageNet models, one on a TPU pod, another on a single \texttt{p3.16xlarge} instance with 8 NVIDIA V100 GPUs,
    and a third on multiple \texttt{p3.16xlarge} instances for selected official \dawnbench entries. As the
    number of workers increases, the scaling performance drops off
    (over 2$\times$ gap from ideal scaling).
  }
  \label{fig:speedup-vs-num-workers}
  \vspace{-0.5em}
\end{figure*}

\begin{figure*}
\centering
\begin{subfigure}{0.65\columnwidth}
  \includegraphics[width=1.0\columnwidth]{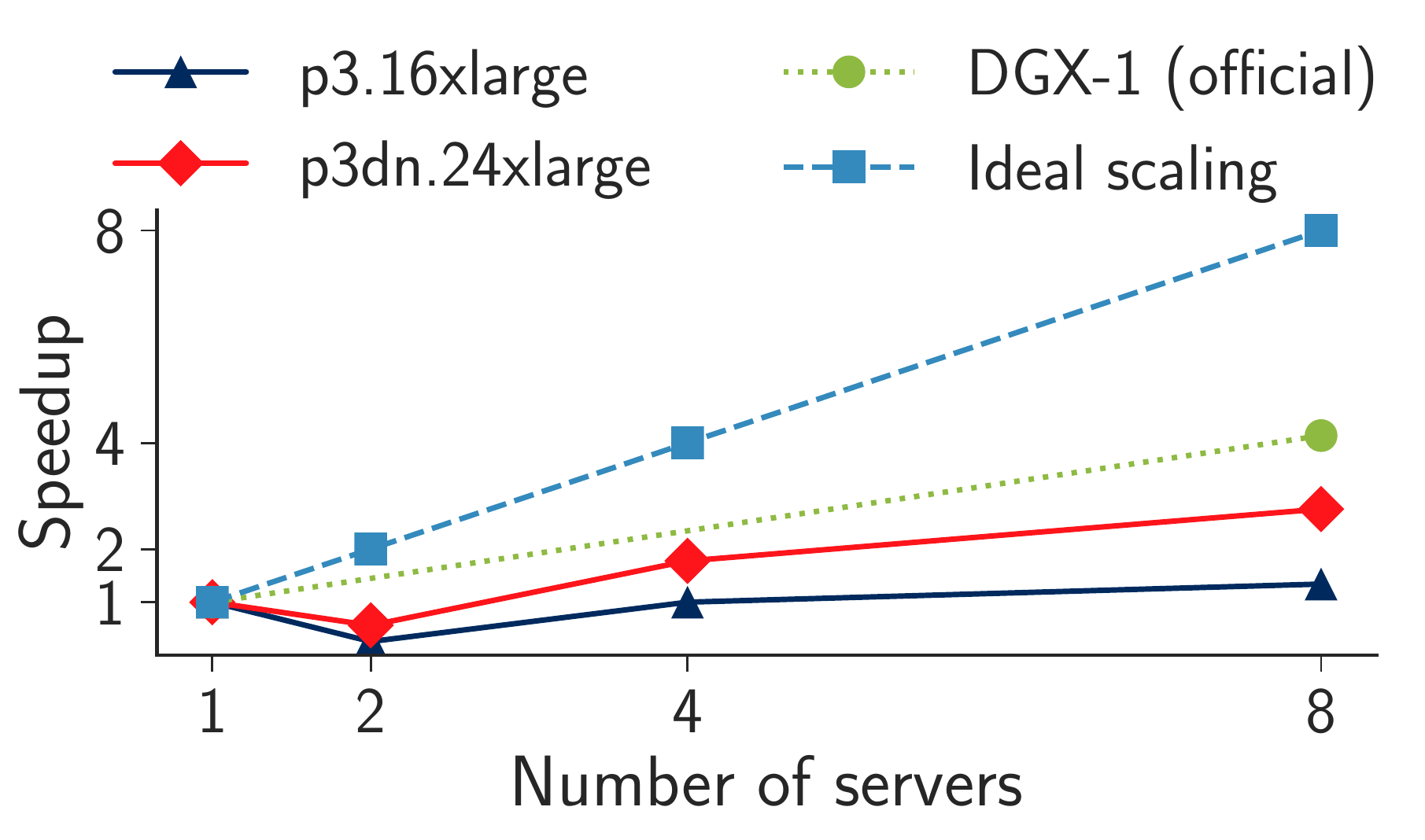}
  \caption{SSD across 8-V100 servers.}
  \label{fig:ssd-inter-scaling-speedup}
\end{subfigure}
\begin{subfigure}{0.65\columnwidth}
  \includegraphics[width=1.0\columnwidth]{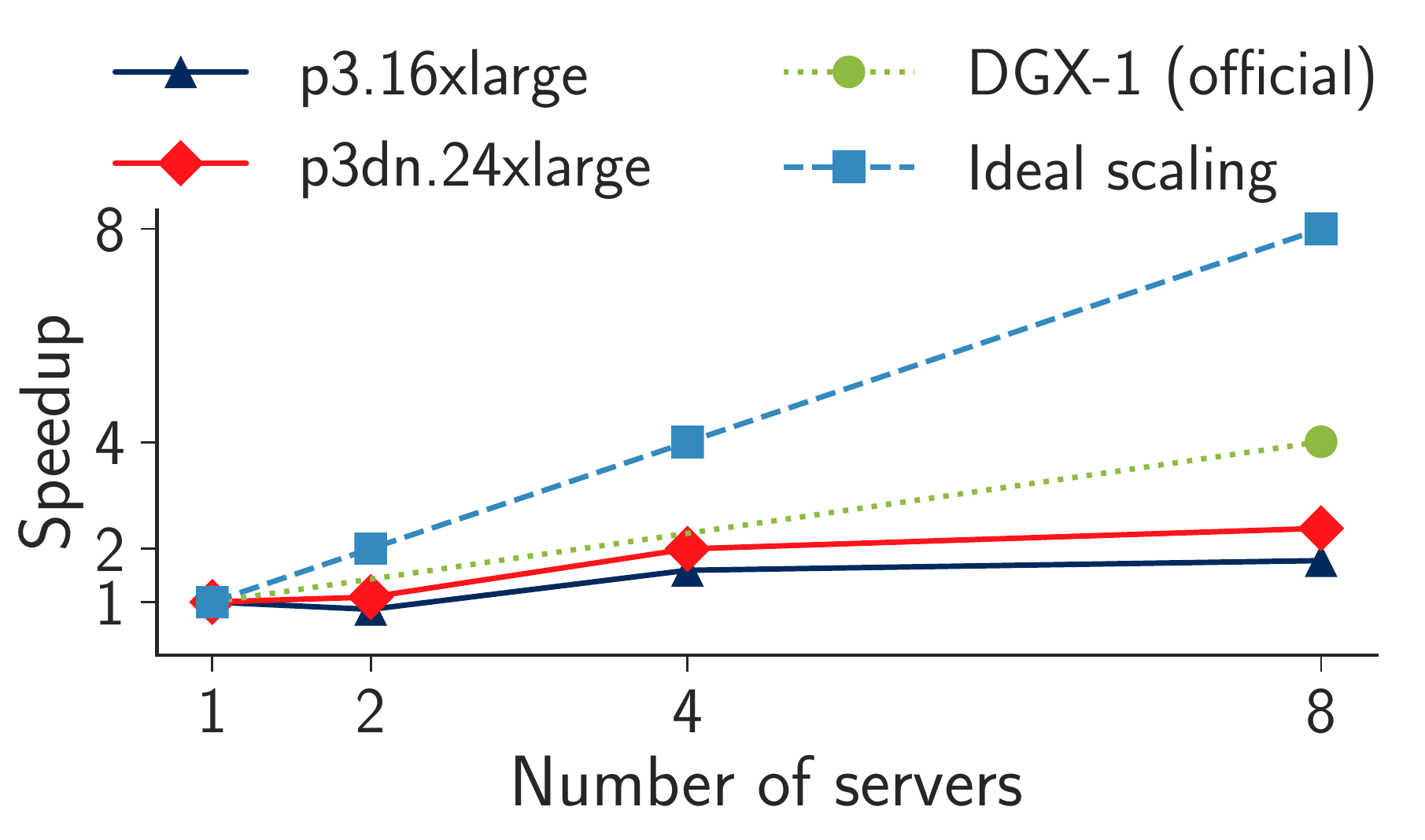}
  \caption{Mask R-CNN across 8-V100 servers.}
  \label{fig:mask-inter-scaling-speedup}
\end{subfigure}
\begin{subfigure}{0.65\columnwidth}
  \includegraphics[width=1.0\columnwidth]{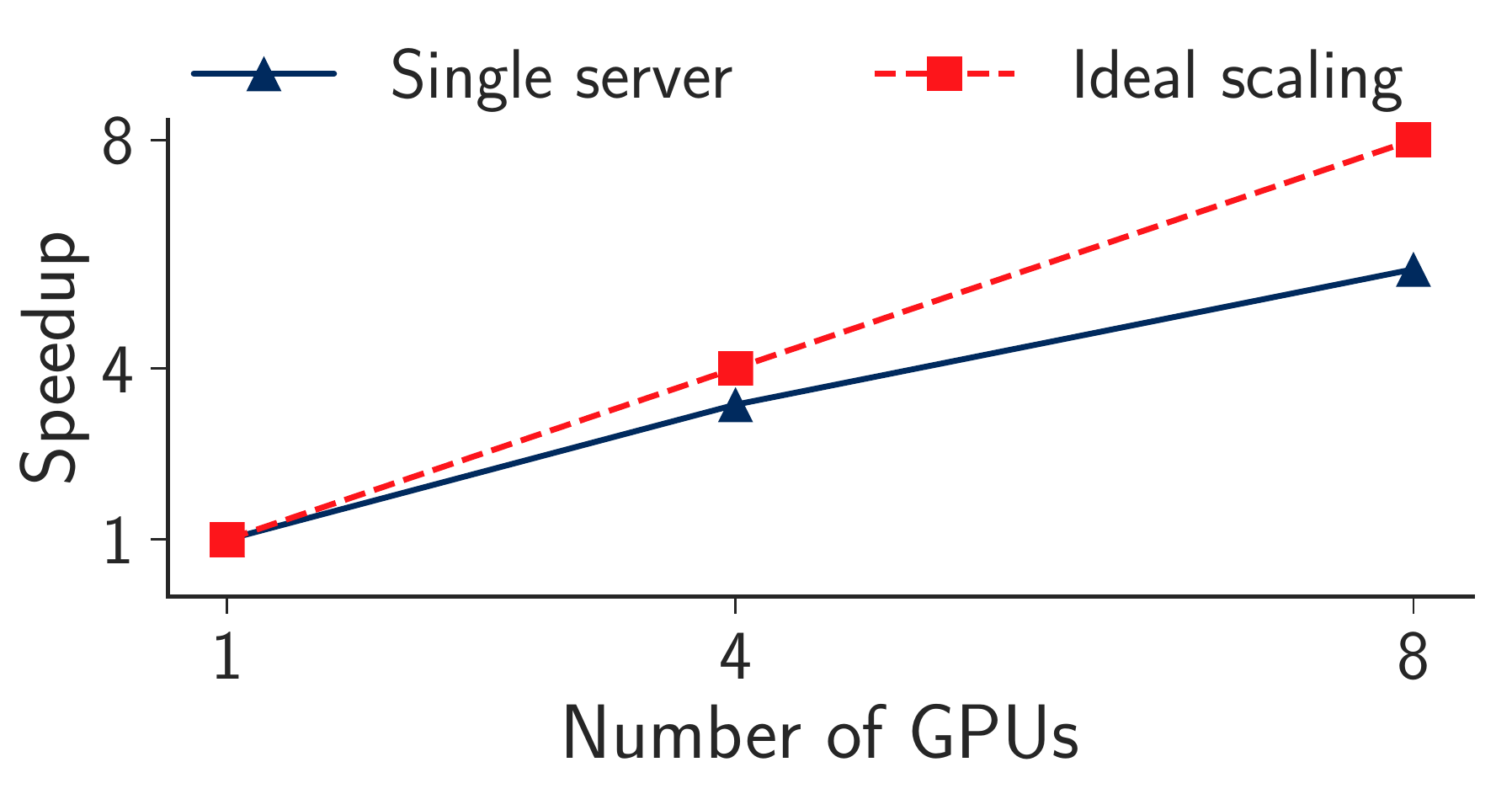}
  \caption{Mask R-CNN within 8-V100 server.}
  \label{fig:mlperf-tta-mask-intra-scaling}
\end{subfigure}
\caption{Speedup of TTA with respect to a single worker vs. number of workers for an
    SSD model on multiple 8-V100 servers (\texttt{p3.16xlarge} and \texttt{p3dn.24xlarge}
    instances on AWS, and a NVIDIA DGX-1 server in an on-premise deployment), a Mask R-CNN model on
    multiple 8-V100 servers, and a Mask R-CNN model within a
    \texttt{p3.16xlarge} instance.}
\label{fig:mlperf-tta-inter-scaling}
\end{figure*}

\subsection{Scaling of Distributed Training}
\label{sec:distributed-training}

With an increase in model size and complexity, distributed training has become
imperative to train models in reasonable timeframes. However, distributed training
requires expensive cross-accelerator communication~\cite{jia2018beyond}. To better quantify these
communication overheads, we trained the same models with different accelerator
counts, and studied the scaling behavior of end-to-end training.

\minihead{Scaling of Time-to-Accuracy}
To scale up to hundreds of accelerators, every large-scale \dawnbench and \mlperf entry
used large minibatches to saturate the available hardware. This includes the machine translation and object detection tasks, even though the original large minibatch training technique was only tested on the ResNet-50 image classification model~\cite{goyal2017accurate}.
Table~\ref{table:tta-vs-throughput} shows batch sizes and
throughputs of various \mlperf official entries. As shown, the batch size can be
scaled from 4 to nearly 50$\times$ the base batch size.

We find that both time-per-epoch and TTA scale almost linearly
with the number of workers \emph{within} a server, across a range of models
for image classification, object detection, and language translation
in both the \dawnbench and \mlperf benchmarks (Figures~\ref{fig:speedup-vs-num-workers-resnet50}
and~\ref{fig:mlperf-tta-mask-intra-scaling}).

However, we found that both time-per-epoch and TTA do not scale as
well for training that spans multiple servers. In Figure~\ref{fig:speedup-vs-num-workers-amoebanet},
we show the speedup relative to one worker of per-epoch time for an AmoebaNet model
trained in a TPU Pod with 64 TPUs on the ImageNet dataset.
Figure~\ref{fig:speedup-vs-num-workers-resnet50-dgx1} shows the speedups when
scaling ResNet-50 training up to 16 \texttt{p3.16xlarge} instances (each server
has 8 NVIDIA V100 GPUs) on Amazon Web Services (AWS).
Time-per-epoch shows as much as a 38.9\% gap from linear scaling.
Time-to-accuracy scales even worse, since a greater number of epochs are needed
to converge to the same accuracy target for the larger minibatch size.
We see similar results for the SSD and Mask R-CNN models using both
\texttt{p3.16xlarge} instances on AWS, and DGX-1 servers (NVIDIA's
optimized server with 8 V100 GPUs) in a private
cloud deployment with Infiniband network communication in Figures~\ref{fig:ssd-inter-scaling-speedup} and~\ref{fig:mask-inter-scaling-speedup} .

\begin{figure*}[t!]
  \centering
  \begin{subfigure}{.65\columnwidth}
    \includegraphics[width=1.0\columnwidth]{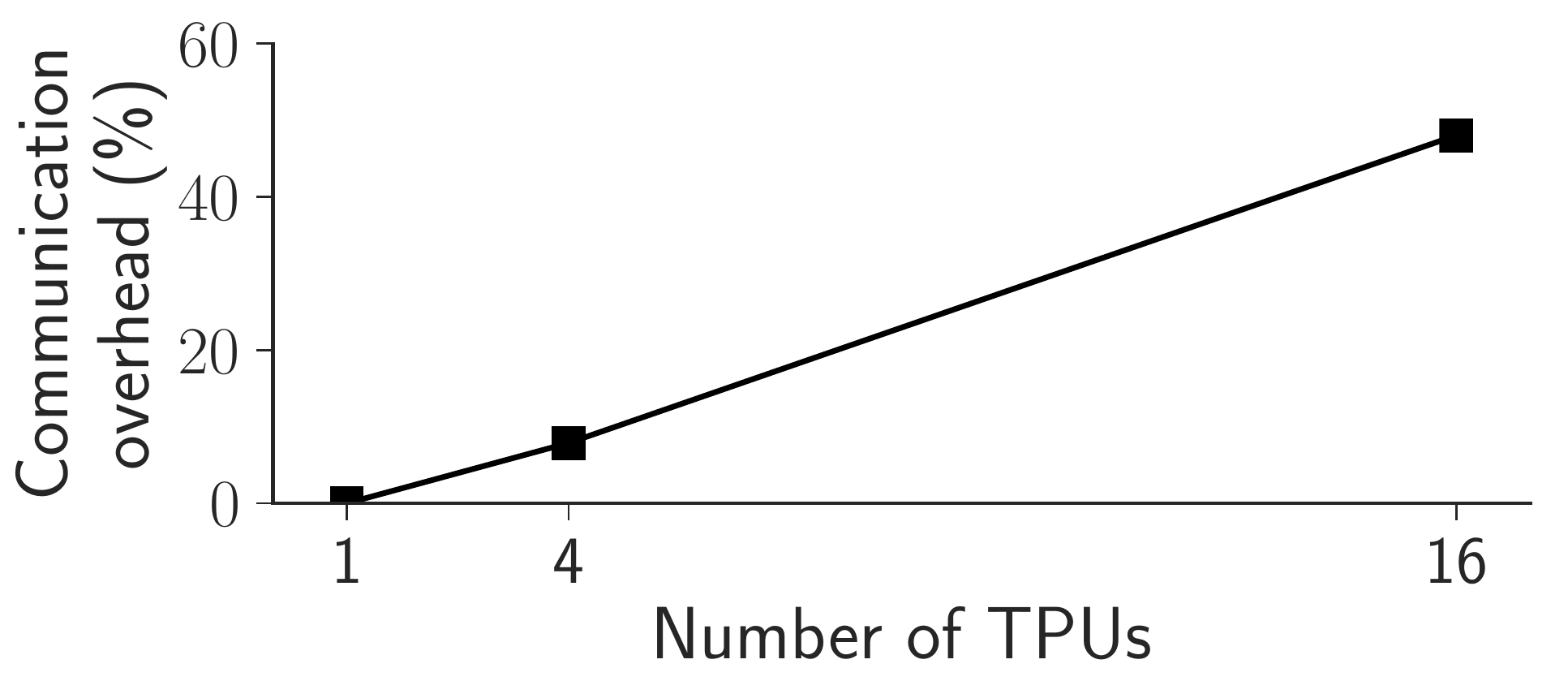}
    \caption{
      AmoebaNet across TPUs, TPU pod.
    }
    \label{fig:communication-overhead-vs-num-workers-amoebanet}
  \end{subfigure}
  \begin{subfigure}{.65\columnwidth}
    \includegraphics[width=1.0\columnwidth]{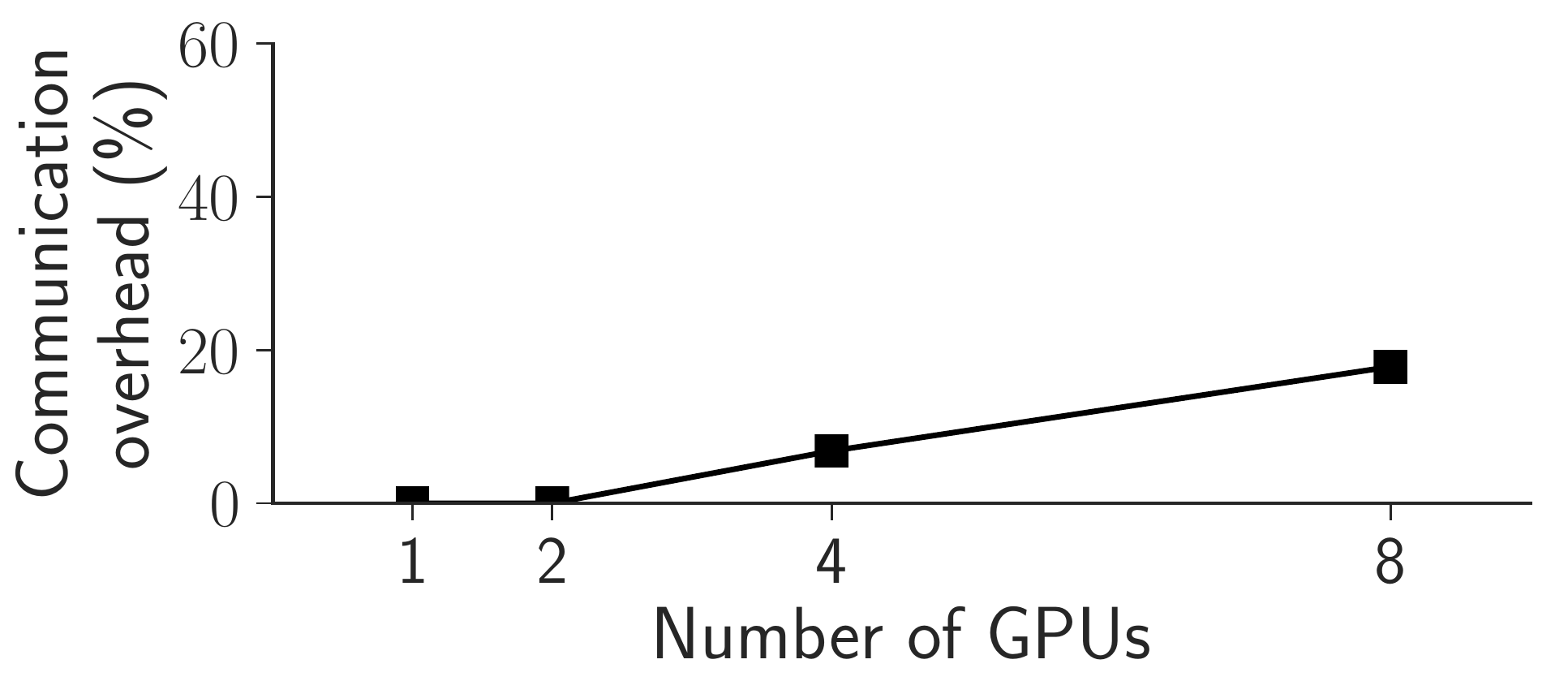}
    \caption{
      ResNet-50 within \texttt{p3.16xlarge} server.
    }
    \label{fig:communication-overhead-vs-num-workers-resnet50}
  \end{subfigure}
  \begin{subfigure}{.65\columnwidth}
    \includegraphics[width=1.0\columnwidth]{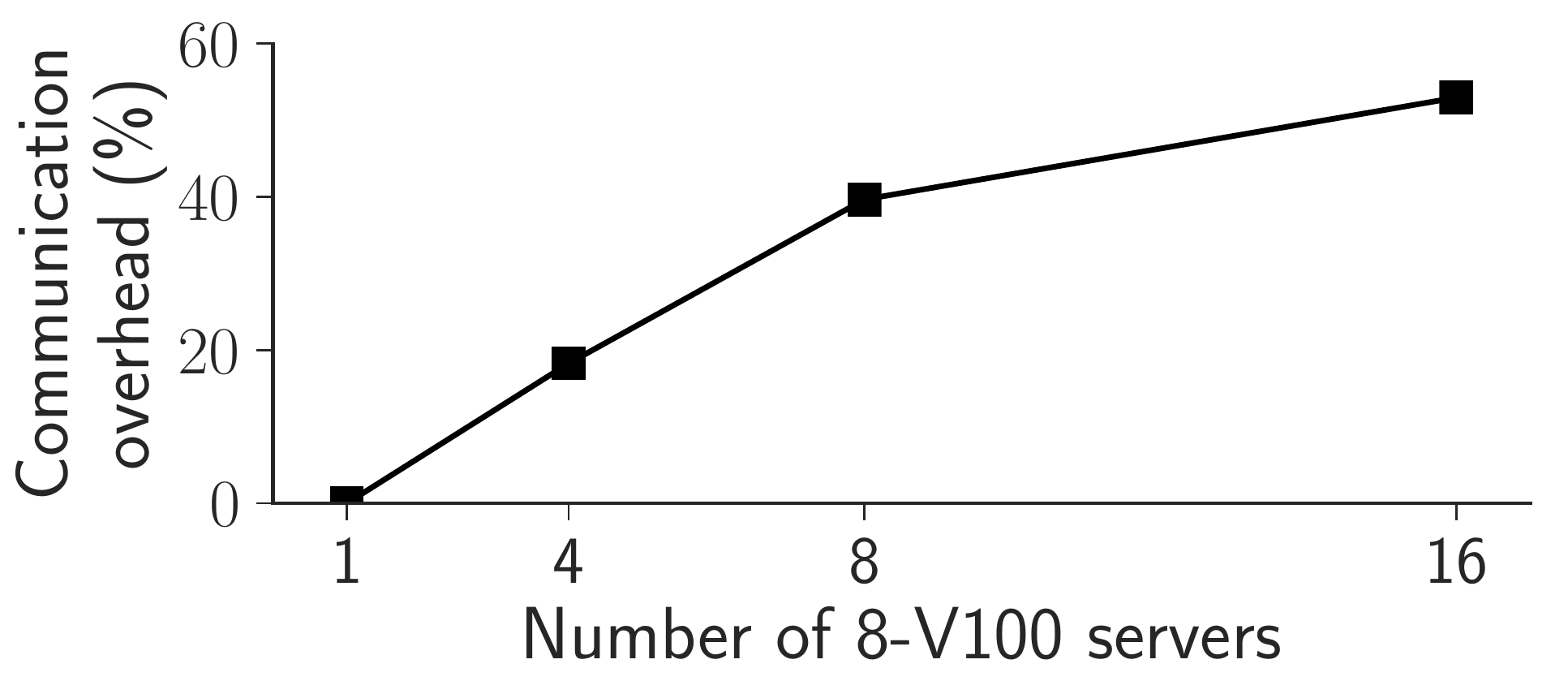}
    \caption{
      ResNet-50 across \texttt{p3.16xlarge} servers.
    }
    \label{fig:communication-overhead-vs-num-workers-resnet50-dgx1}
  \end{subfigure}
  \caption{
    Percentage of time in an epoch spent communicating vs. number of workers for three
    ImageNet models, one on a TPU pod, another on a single \texttt{p3.16xlarge} instance,
    and a third on multiple \texttt{p3.16xlarge} instances. Within a 8-V100 server, communication overhead
    is low ($17.82\%$), but cross-machine communication is more expensive ($53\%$).
  }
  \label{fig:communication-overhead-vs-num-workers}
\end{figure*}

\begin{figure*}
\centering
\begin{subfigure}{0.65\columnwidth}
  \includegraphics[width=1.0\columnwidth]{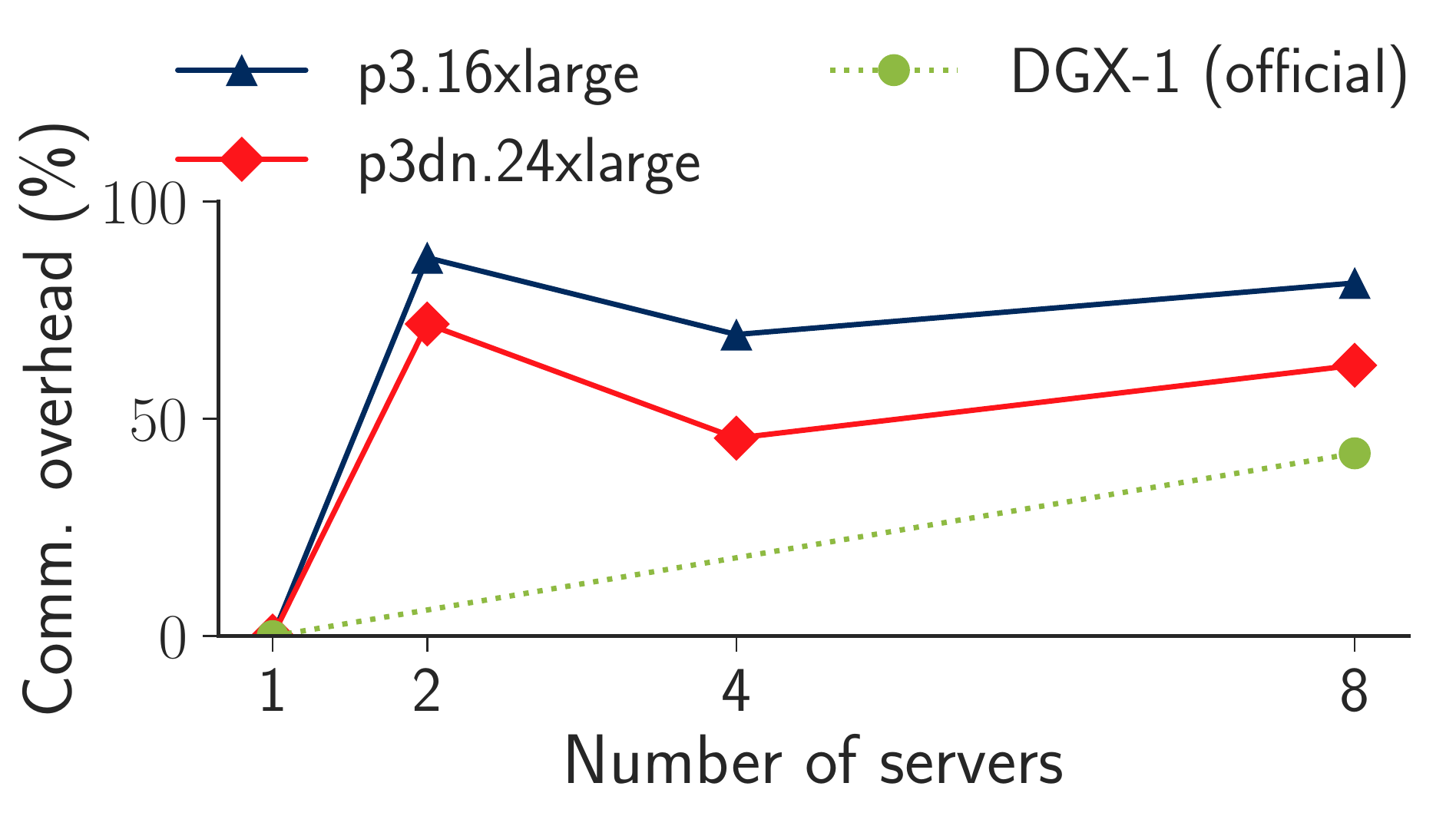}
  \caption{SSD across 8-V100 servers.}
\end{subfigure}
\begin{subfigure}{0.65\columnwidth}
  \includegraphics[width=1.0\columnwidth]{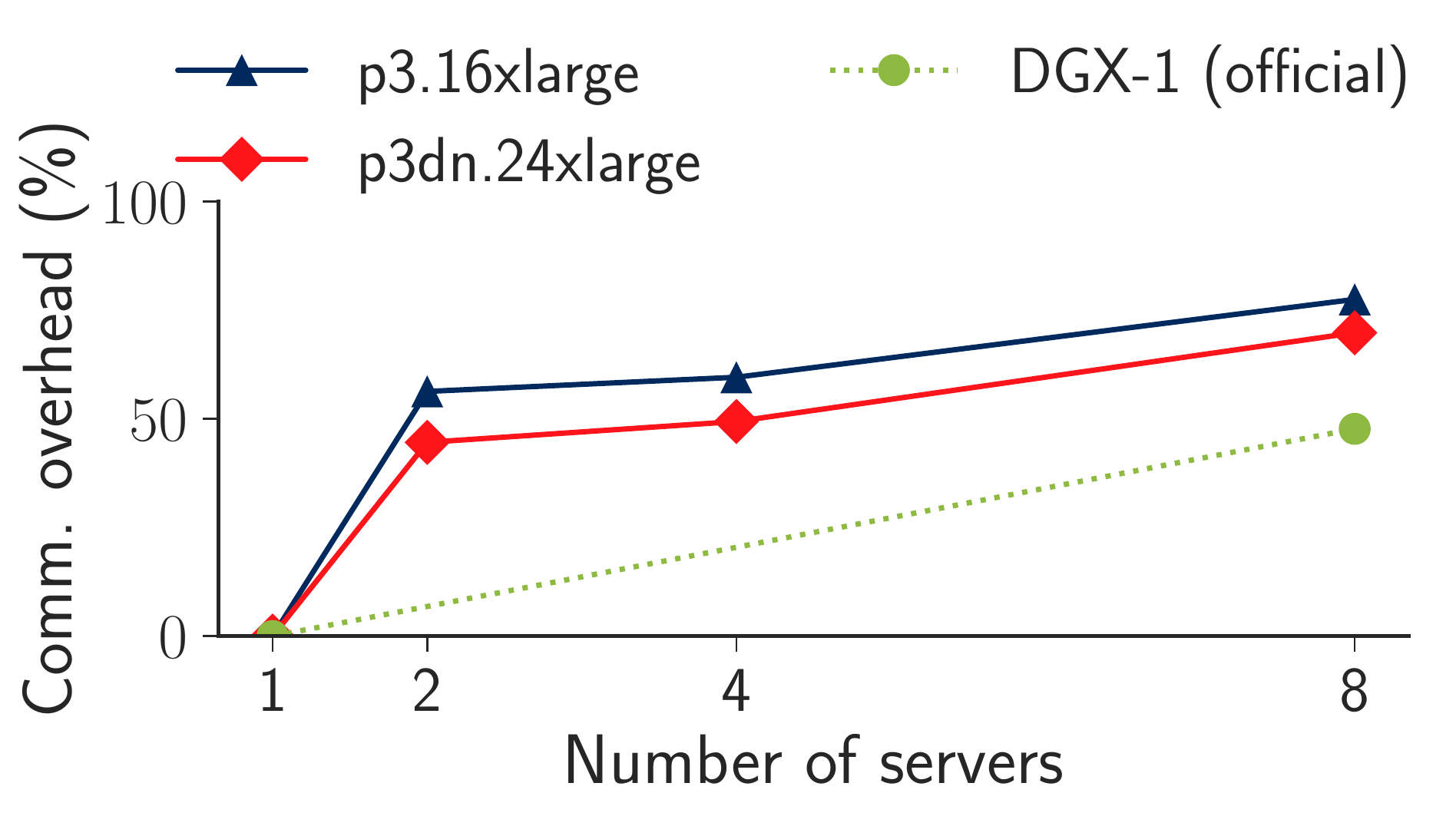}
  \caption{Mask R-CNN across 8-V100 servers.}
\end{subfigure}
\caption{Percentage of time per epoch spent communicating vs. number of workers
  for various \mlperf entries in
  both on-premise and public cloud deployments. Communication overheads are as high as
  47.67\% in on-premise deployments and up to 77.46\% in the public cloud.}
\label{fig:mlperf-comm-overhead}
\end{figure*}

\minihead{Communication Overhead}
To further understand the impact of networking on distributed training, we computed the communication
overheads of both official \dawnbench and \mlperf entries.
These results are shown in
Figures~\ref{fig:communication-overhead-vs-num-workers}
and~\ref{fig:mlperf-comm-overhead}, and
Tables~\ref{table:communication-overhead-mlperf-optimized}
and~\ref{table:communication-overhead-intra-server}. For the \mlperf entries,
we show communication overheads for both the official entries run on private
on-premise deployments, and reproduced entries run on public cloud
deployments to quantify the impact of optimized network interconnects like
Infiniband and RDMA on end-to-end training time.
For the \dawnbench entries, we show communication overhead numbers
on the public cloud.

\begin{table}[t!]
\centering
\setlength\itemsep{2em}
\small
\begin{tabular}{lccc}
  Model & Machine Config. & Comm. Overhead (\%) \\ \hline \hline
  ResNet-50 & 80xDGX-1 & $64.947\%$ \\
  ResNet-50 & 16xDGX-2H & $25.859\%$ \\
  \hline
  SSD & 8xDGX-1 & $42.043\%$ \\
  SSD & 8xDGX-2H & $68.231\%$ \\
  \hline
  Mask R-CNN & 8xDGX-1 & $47.674\%$ \\
  Mask R-CNN & 4xDGX-2H & $42.131\%$ \\
  \hline
  GNMT & 32xDGX-1 & $71.146\%$ \\
  GNMT & 16xDGX-2H & $67.436\%$ \\
  \hline
  Transformer & 24xDGX-1 & $35.127\%$ \\
\end{tabular}
\vspace{-0.5em}
\caption{Percentage of time in an epoch spent communicating for official optimized distributed \mlperf entries.}
\label{table:communication-overhead-mlperf-optimized}
\end{table}

\begin{table}[t!]
\centering\small
\setlength\itemsep{2em}
\begin{tabular}{lccc}
  Model & Machine Config. & Comm. Overhead (\%) \\ \hline \hline
  ResNet-50 & 4xV100 (AWS) & $4.528\%$ \\
  ResNet-50 & 8xV100 (AWS) & $13.400\%$ \\
  \hline
  SSD & 4xV100 (AWS) & $5.364\%$ \\
  SSD & 8xV100 (AWS) & $14.999\%$ \\
  \hline
  Mask R-CNN & 4xV100 (AWS) & $17.167\%$ \\
  Mask R-CNN & 8xV100 (AWS) & $26.163\%$ \\
  \hline
  GNMT & 4xV100 (AWS) & $9.921\%$ \\
  GNMT & 8xV100 (AWS) & $15.832\%$ \\
  \hline
  Transformer & 4xV100 (AWS) & $26.692\%$ \\
  Transformer & 8xV100 (AWS) & $15.546\%$ \\
\end{tabular}
\vspace{-0.5em}
\caption{Percentage of time in an epoch spent communicating for reproduced single-server \mlperf entries.}
\label{table:communication-overhead-intra-server}
\end{table}

As shown, communication remains a \emph{significant} overhead. Even on
on-premise deployments with 100Gb/s InfiniBand EDR interconnect can have
communication overheads as high as 71.15\%, for the GNMT model using 32 DGX-1
servers. This overhead can rise to 77.46\% when using Amazon's \texttt{p3.16xlarge}
instances with a 25 Gigabits/second interconnect for Mask R-CNN.

Figure~\ref{fig:communication-overhead-vs-num-workers} shows the communication
overhead as the number of workers increase for \dawnbench entries; we see that within a DGX-1, communication
overhead is much lower (17.82\%) compared to across servers (53\%), since NVIDIA's
\texttt{nvlink} interconnect has far higher bandwidth than the 25Gbps provided by Amazon
EC2 across \texttt{p3.16xlarge} instances.
As \dawnbench did not have entries for AmoebaNet on GPUs, we were unable to make
a completely fair apples-to-apples comparison of the scaling properties between
AmoebaNet and ResNet-50.

\paragraph{Discussion.}
These results suggest that despite the work in scaling ML training to
many multi-GPU servers~\cite{goyal2017accurate, smith2017learningratedecay},
communication remains a bottleneck, for large machine counts and
for certain models in public cloud deployments. The work by Goyal et al.~\cite{goyal2017accurate} shows far
better scaling than we have observed in the \dawnbench entries; we believe this
is due to the fact that the results presented in this paper used faster V100
GPUs (compared to P100 GPUs), and had slower network interfaces (up to 25
Gigabits/second on AWS compared to 50 Gigabits/second in a private Facebook
cluster).

To address this, highly optimized communication libraries like Horovod~\cite{sergeev2018horovod}
have been developed. Other work~\cite{lin2018deep} has explored techniques
to reduce the amount of data sent over the network.
However, these techniques need to be evaluated
on more models and in more hardware settings for widespread adoption.
Integration into widely-used deep
learning frameworks like PyTorch and TensorFlow would also help with usability.
Additionally, exploring parallelization schemes
other than data parallelism that do not require all-to-all communication among
all workers could be helpful.

\begin{figure}[t!]
  \centering
  \includegraphics[width=0.95\columnwidth]{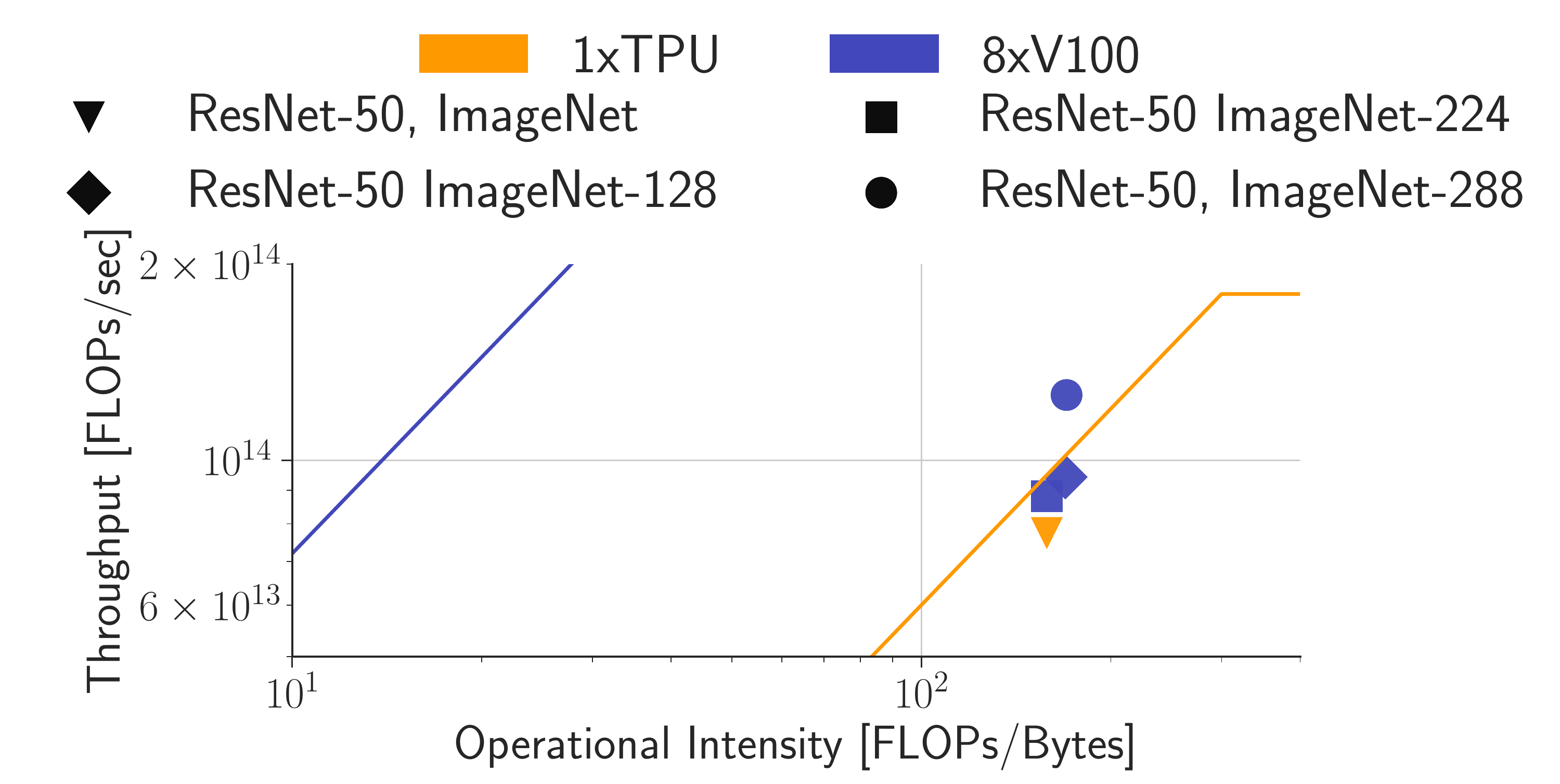}
  \caption{
    Roofline models for the various \dawnbench entries. All of the
    entries under-utilize the hardware resources, by up to 10$\times$.
  }
  \label{fig:roofline-dawnbench}
  \vspace{-0.5em}
\end{figure}

\begin{figure}[t!]
  \centering
  \includegraphics[width=0.9\columnwidth]{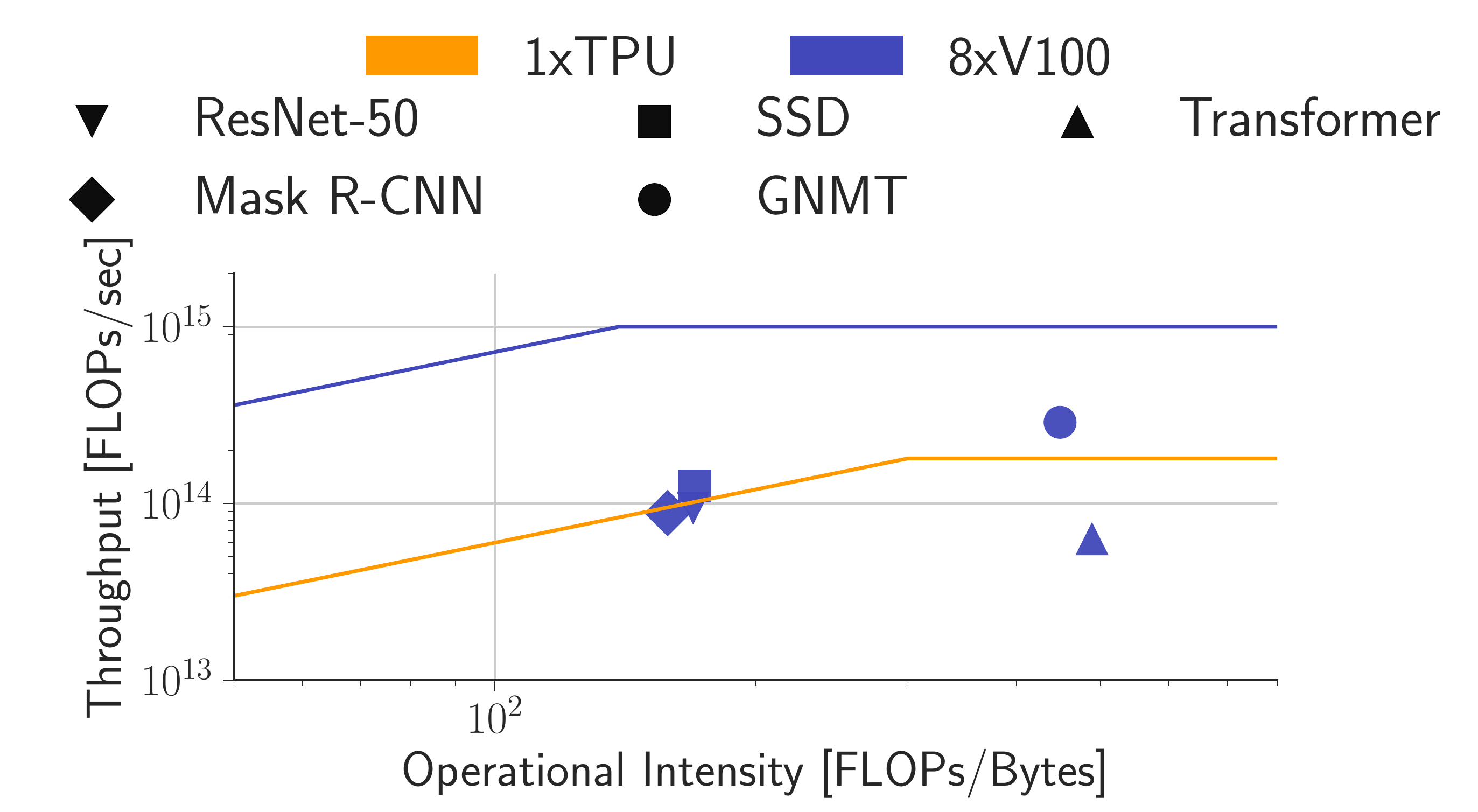}
  \caption{
    Roofline models for the various \mlperf entries. All of the
    entries under-utilize the hardware resources, by up to 10$\times$.
  }
  \label{fig:roofline-mlperf}
  \vspace{-0.5em}
\end{figure}

\subsection{Single-worker Utilization}
\label{sec:hw-single-node}

To study the utilization of the compute units of the accelerators themselves,
we analyzed the performance of some \dawnbench and \mlperf submissions on a
single accelerator, without network overhead.

\minihead{Roofline Analysis}
To understand the hardware performance of single-worker training, we
used the roofline model~\cite{williams2009roofline}, which can highlight causes of performance bottlenecks.
The roofline model plots computational throughput (in floating point operations per second)
against the operational intensity of the application (number of floating-point operations
performed per DRAM byte accessed). Applications with high operational intensity are
``compute-bound'' (the flat line in Figure~\ref{fig:roofline-dawnbench}) and bottlenecked on
the device's computation units, while applications
with low operational intensity are ``memory-bound'' (the slanting line in
Figure~\ref{fig:roofline-dawnbench}) and bottlenecked on memory accesses.

We show results in Figures~\ref{fig:roofline-dawnbench} and
\ref{fig:roofline-mlperf}. Each point in these figures represents a \dawnbench or \mlperf entry.
For entries which used progressive image resizing~\cite{lim2017enhanced,
karras2017progressive}, where different image sizes are used through training,
we show each image size used. Operational intensities and throughputs
are approximated by instrumenting training code and using profiling tools like \texttt{nvprof}.

As shown, all entries analyzed \emph{severely} underutilize the available compute resources -- each
plotted point achieves a throughput significantly lower than peak device
throughput.

\begin{figure}[t!]
  \centering
  \includegraphics[width=0.84\columnwidth]{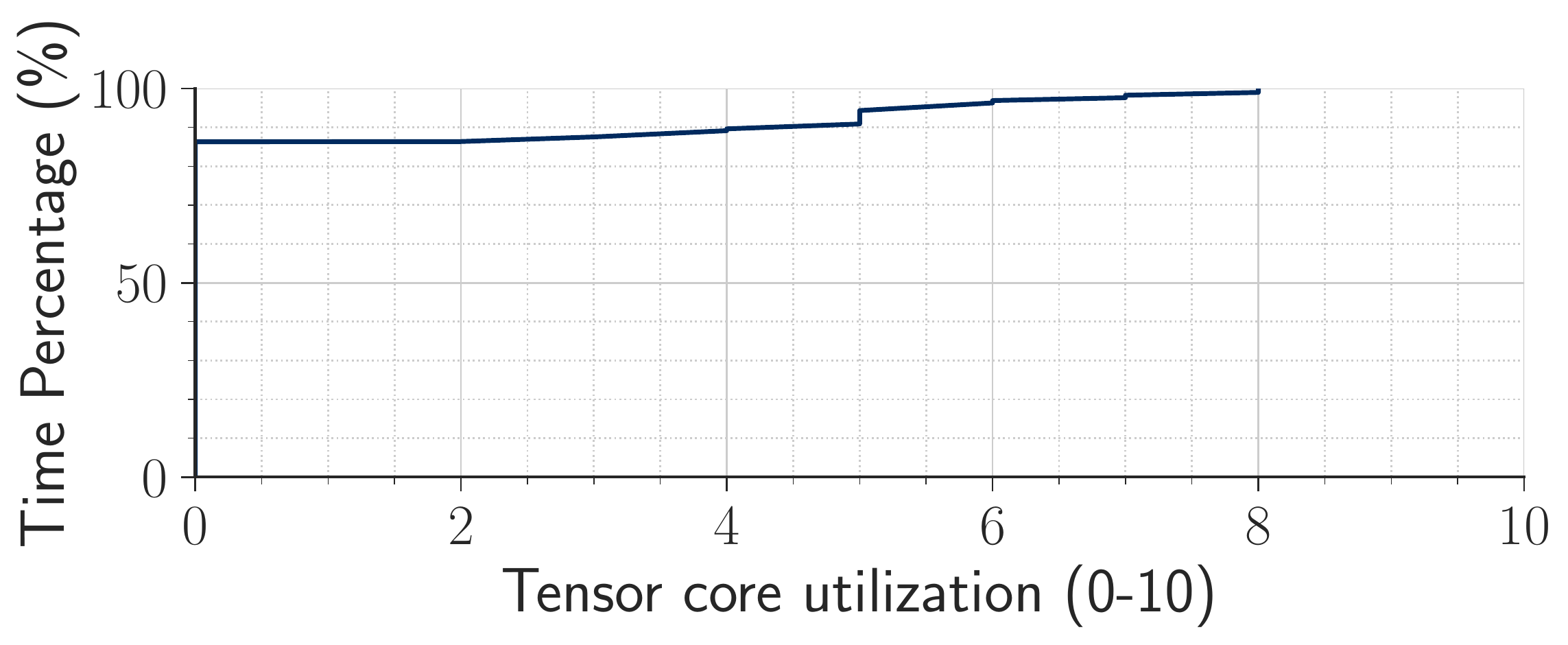}
  \caption{CDF of tensor core utilization for the \texttt{fast.ai} ResNet50 model trained with \texttt{fp16}
  precision submitted to the \dawnbench competition. About $85\%$ of time is spent on kernels that don't utilize the NVIDIA
  Tensor Cores \emph{at all}, and no kernel achieves full utilization of the Tensor Core units.}
  \label{fig:tensorcore-utilization}
\end{figure}

\begin{figure}[t!]
  \centering
  \includegraphics[width=0.84\columnwidth]{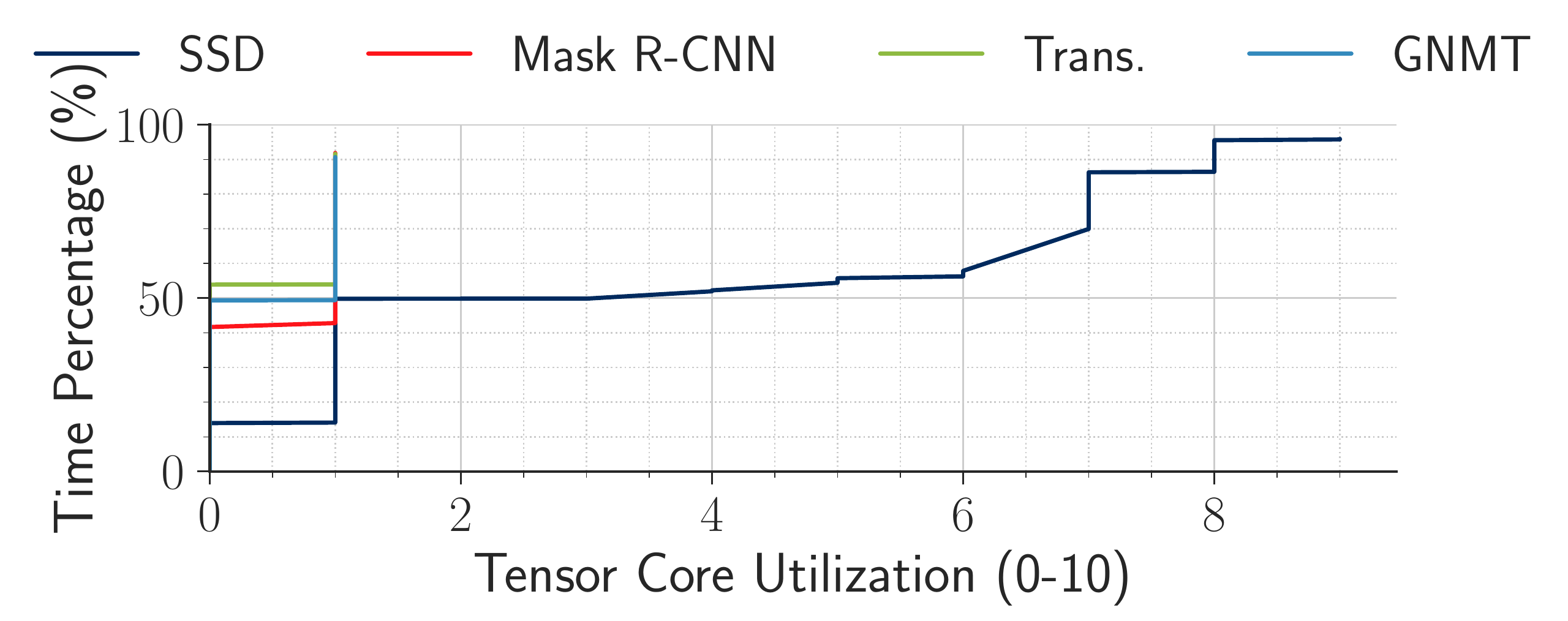}
  \caption{CDF of tensor core utilization for different \mlperf models trained with
  \texttt{fp16} precision.}
  \label{fig:mlperf-tensorcore-utilization}
\end{figure}

\begin{figure}[t!]
  \centering
  \includegraphics[width=0.84\columnwidth]{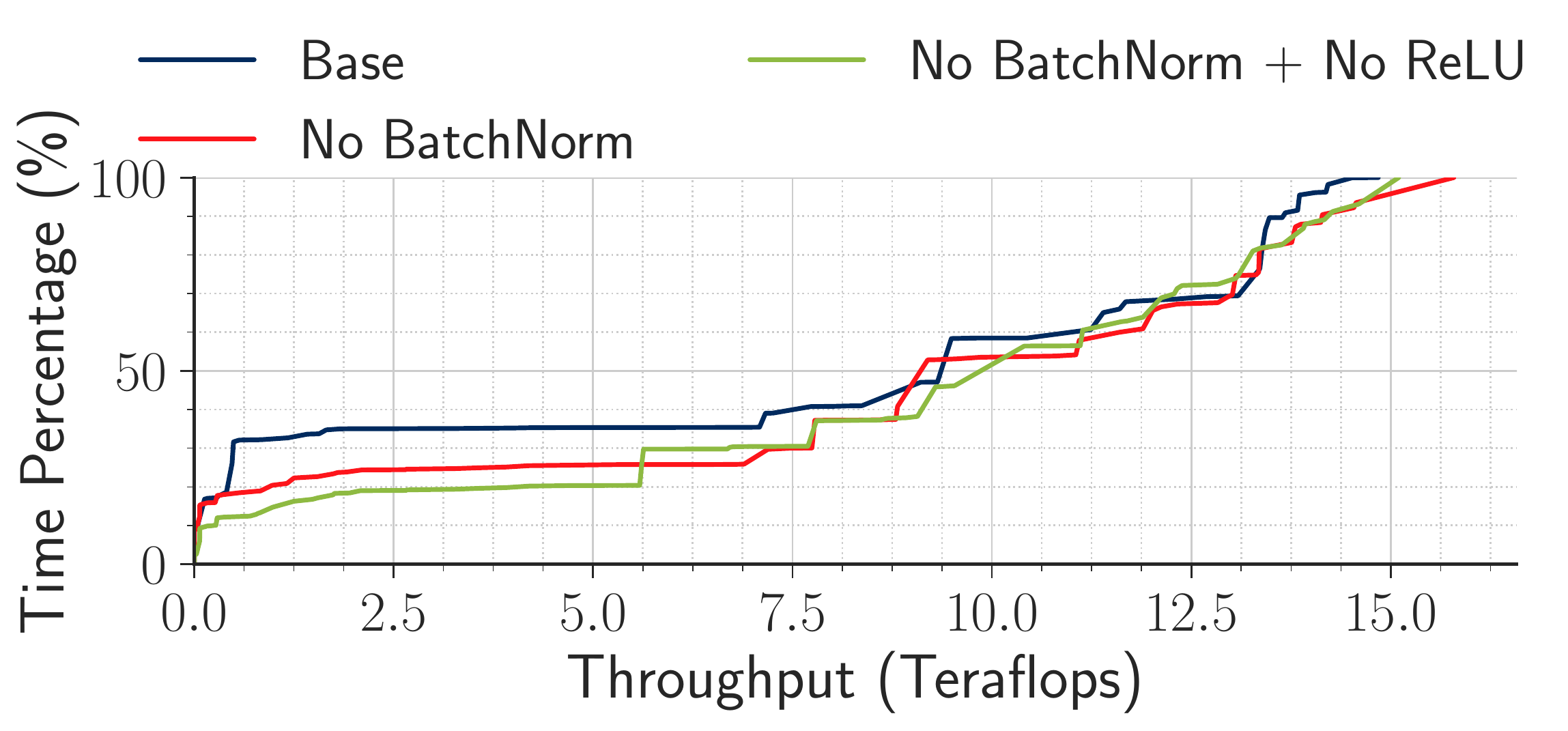}
  \caption{CDF of per-kernel throughput for ResNet50 models trained with \texttt{fp32}
  precision. The CDF is computed by percentage of time spent executing each GPU kernel.
  A standard ResNet-50 model spends about $40\%$ time in low-throughput kernels ($< 6$ Teraflops).
  Removing the BatchNorm layer from the ResNet50 model decreases the percentage of
  time in low-throughput kernels to about $30\%$; removing the ReLU layers
  decreases this further.}
  \label{fig:resnet50-perf}
\end{figure}

\minihead{Bottlenecks in Training}
To investigate the source of underutilization on the V100 GPU, we measured the
\texttt{fp32} throughput and Tensor Core utilization of each GPU kernel in
PyTorch's implementation of ResNet-50.  The V100s have peak throughputs of 15.7
Teraflops of \texttt{fp32} arithmetic and 125 Teraflops of half-precision
arithmetic via Tensor Cores~\cite{markidis2018nvidia}.

Figures~\ref{fig:tensorcore-utilization} and \ref{fig:mlperf-tensorcore-utilization} show that the GPU kernels taking the majority
of time when using \texttt{fp16} precision utilize the Tensor Cores poorly, with a time-averaged Tensor Core
utilization of 0.71 (on a scale of 1-10 as reported by \texttt{nvprof}).

Training the same model even with standard \texttt{fp32} precision only achieves
a throughput of 7.6 Teraflops, compared to peak device throughput
of 15.7 Teraflops. This is largely due to memory-bound
kernels like BatchNorm~\cite{ioffe2015batch} and ReLU~\cite{nair2010rectified},
that take a significant percentage of total runtime.
This is illustrated in Figure~\ref{fig:resnet50-perf},
which show that a non-trivial portion of kernels underutilize the GPU, and
that removing BatchNorm and ReLU layers improves \texttt{fp32} throughput by
about $20\%$.

\paragraph{Discussion.}
Compilers for deep learning like TVM~\cite{tvm} and XLA~\cite{xla}
try to automatically generate code given
a higher-level description of the deep learning computation being performed.
Optimizations like loop and kernel fusion can help reduce the impact of
memory-bound kernels by reducing the number of DRAM reads and writes made.
For example, consider code that performs the forward pass of a BatchNorm
followed by the forward pass of a ReLU. Naively,
this code would be executed by code that resembles the following,
\lstset{basicstyle=\scriptsize\ttfamily,breaklines=true,emph={max},
        emphstyle=\bfseries,numbers=left, xleftmargin=2em}
\begin{lstlisting}[language=C]
// BatchNorm.
for (int i = 0; i < n; i++) {
    y[i] = gamma * ((x[i] - mu) / sigma) + beta;
}
// ReLU.
for (int i = 0; i < n; i++) {
    z[i] = max(y[i], 0);
}
\end{lstlisting}
\lstset{basicstyle=\footnotesize\ttfamily,breaklines=true,emph={max},
        emphstyle=\bfseries,numbers=left, xleftmargin=2em}

In the above listing, a DRAM write is performed for \lstinline$y[i]$
and \lstinline$z[i]$, and a DRAM read is performed to compute \lstinline$z[i]$.

However, for training, we could optimize the above code by fusing the two loops,
saving on DRAM reads of \lstinline$y[i]$ since the intermediate result is
written to a local variable instead.

In addition, we believe that co-design of model architectures with modern hardware
could be useful as well. For example, as we have shown, the BatchNorm and ReLU
operations are memory bound. It may be possible to develop alternatives to these
operations that are less memory-bound, but provide similar statistical
effects, resulting in faster training.

\section{Related Work}
\label{sec:related-work}

\minihead{Benchmarking DL Training}
Many prior ML benchmarks use throughput (either per-kernel or
per-iteration) as a metric~\cite{bahrampour2015comparative, baidu2017deepbench,
chintala2017convnet, adolf2016fathom, google2017benchmarks,
shi2016benchmarking}. While throughput can inform the development of ML
algorithms and systems, we show throughput alone cannot fully characterize ML
systems.

Several ML benchmarks have done static workload characterizations on systems
that do not contain state-of-the-art hardware with FP16
support~\cite{adolf2016fathom, zhu2018tbd}. Furthermore, several
benchmarks, including Fathom, do not benchmark distributed DL
training~\cite{baidu2017deepbench, chintala2017convnet, adolf2016fathom}.
TBD~\cite{zhu2018tbd} benchmarks distributed training on older accelerators that
do not contain FP16 support, which significantly changes the proportion of total
runtime spent on computation and communication.
In contrast to prior work, we analyze code that
has been optimized by teams of engineers on state-of-the-art hardware.
We additionally analyze distributed DL systems that uses this hardware.
We show that Tensor Cores can be severely underutilized and that communication overheads are as high as 71\%, even in
optimized on-premise deployments.

\minihead{Benchmarking High Performance Computing Systems}
Researchers have developed many methods for benchmarking computer systems and
HPC systems~\cite{ghose2018your, atikoglu2012workload, chang2017understanding}.
The majority of these systems measure deterministic workloads (e.g., DRAM,
key-value stores), but measuring DL systems requires a more nuanced analysis
to reason about both runtime and the generalizability of the final model.
While these systems could be used to improve individual components of DL
training systems (e.g., faster convolution algorithms), they are not sufficient
to measure end-to-end DL training.

\minihead{High Performance DL}
Researchers have developed many optimizations for high performance DL
training~\cite{li2014scaling, you2018imagenet, akiba2017extremely,
goyal2017accurate}. Unfortunately, many such optimizations are closed-source.
To the best of our knowledge, \dawnbench and \mlperf are the first open-entrant
benchmarks with open-source entries for optimizing TTA on a range
of tasks. We take advantage of the open-source code to study TTA and analyze these workloads.

Some work on high performance DL~\cite{you2018imagenet,
akiba2017extremely, goyal2017accurate} used TTA.
However, these systems largely used TTA as a metric to optimize, but
do not study the metric in detail. In this work, we analyze TTA as
a metric and show that it is largely stable and models optimized for TTA
generalize well.

\section{Conclusion}
In this paper, we perform the first in-depth analysis of \dawnbench entries to
investigate the behavior of TTA as a metric and trends in the best-performing entries.
We corroborate our results by analyzing entries from \mlperf v0.5, which also adopted TTA.
Both benchmarks received professionally optimized entries from leading industry groups, creating one of the first opportunities to
study ML systems optimized heavily for training performance.
We find that TTA is usually stable to the randomness in ML training with a low coefficient of variation ($<14\%$)
across image classification, machine translation, and object detection.
We also find that models optimized for
TTA generalize nearly as well as unoptimized models. Finally,
we find that entries highly optimized for TTA still underutilize
available hardware, leaving significant room for further improvement.

\section*{Acknowledgments}

We thank Jeremy Howard, the Google Cloud TPU team (including Sourabh Bajaj, Frank Chen,
Brennan Saeta, and Chris Ying), and the many other teams that submitted to \dawnbench. We thank
Juan Manuel Camacho, Shoumik Palkar, Kexin Rong, Keshav Santhanam, Sahaana Suri, Pratiksha
Thaker, and James Thomas for their assistance in labeling. We also thank Amazon and Google for
cloud credits.
This research was supported in part by affiliate members and other supporters of the Stanford DAWN project—Ant Financial, Facebook, Google, Intel, Microsoft, NEC, SAP, Teradata, and VMware—as well as Toyota Research Institute, Keysight Technologies, Northrop Grumman, Hitachi, Cisco, and the NSF under grants DGE-1656518, DGE-114747, and CNS-1651570.

\bibliographystyle{plain}
{\footnotesize \bibliography{sigops-osr-dawnbench}}

\end{document}